%% file: main.tex
\pdfoutput=1
\documentclass[10pt]{article} 
\usepackage[accepted]{tmlr}

\input{math_commands.tex}

\usepackage{hyperref}
\usepackage{url}

\usepackage{hyperref}
\usepackage{url}

\usepackage[utf8]{inputenc} 
\usepackage[T1]{fontenc}    
\usepackage{hyperref}       
\usepackage{url}            
\usepackage{booktabs}       
\usepackage{amsfonts}       
\usepackage{nicefrac}       
\usepackage{microtype}      
\usepackage[dvipsnames]{xcolor}

\usepackage[table]{xcolor}

\usepackage{wrapfig}
\usepackage{amssymb}
\usepackage{mathtools}
\usepackage{amsthm}
\usepackage{graphicx}
\usepackage{hyperref}
\usepackage{url}
\usepackage[utf8]{inputenc} 
\usepackage[T1]{fontenc}    
\usepackage{nicefrac}       
\usepackage{multirow}
\usepackage{subcaption}
\usepackage{xspace}
\usepackage{amsmath}
\usepackage{svg}
\usepackage{bbm}
\usepackage{dsfont}
\usepackage[capitalize,noabbrev]{cleveref}

\newcommand{\Wbf}{\mathbf{W}}

\newcommand{\ie}{\textit{i.e.\xspace}}
\newcommand{\eg}{\textit{e.g.\xspace}}

\newcommand{\Dpre}{D_\text{pre}}
\newcommand{\Ddown}{D_\text{FT}}

\title{Large Pretraining Datasets Don't Guarantee Robustness after Fine-Tuning in Image Classification
}

\author{\name Jaedong Hwang \email jdhwang@mit.edu \\
      \addr Massachusetts Institute of Technology
           \AND
      \name Brian Cheung \email bcheung@ucsf.edu \\
      \addr Massachusetts Institute of Technology \& University of California, San Francisco
      \AND
      \name Zhang-Wei Hong \email zwhong@mit.edu \\
      \addr Massachusetts Institute of Technology
      \AND
      \name Akhilan Boopathy \email akhilan@mit.edu \\
      \addr Massachusetts Institute of Technology
      \AND
      \name Pulkit Agrawal \email pulkitag@mit.edu \\
      \addr Massachusetts Institute of Technology
      \AND
      \name Ila Fiete \email fiete@mit.edu \\
      \addr Massachusetts Institute of Technology
      }

\def\month{09}  
\def\year{2026} 
\def\openreview{\url{https://openreview.net/forum?id=VyVkIucjWU}} 

\begin{document}

\maketitle

\begin{abstract}
\input{sections/0_abstract}
\end{abstract}

\input{sections/1_introduction}
\input{sections/2_related}
\input{sections/3_method}
\input{sections/4_exp}

\input{sections/5_discussion}

\bibliography{references}
\bibliographystyle{tmlr}

\appendix

\input{sections/6_appendix}

\clearpage
{\Large \textbf{Supplementary Materials}}
\input{sections/7_supp}
\end{document}

%% file: math_commands.tex
\usepackage{amsmath,amsfonts,bm}

\def\eqref#1{equation~\ref{#1}}

\def\1{\bm{1}}

\DeclareMathAlphabet{\mathsfit}{\encodingdefault}{\sfdefault}{m}{sl}
\SetMathAlphabet{\mathsfit}{bold}{\encodingdefault}{\sfdefault}{bx}{n}



%% file: sections/0_abstract.tex

Large-scale pretrained models are widely leveraged as foundations for learning new specialized tasks via fine-tuning, with the goal of maintaining the general performance of the model while allowing it to gain new skills. A valuable goal for all such models is robustness: the ability to perform well on out-of-distribution (OOD) tasks. We assess whether fine-tuning preserves the overall robustness of the pretrained model in image classification, and observed that models pretrained on large datasets exhibited strong catastrophic forgetting and loss of OOD generalization. To systematically assess robustness preservation in fine-tuned models, we propose the Robustness Inheritance Benchmark (ImageNet-RIB). The benchmark, which can be applied to any pretrained model, consists of a set of related but distinct OOD (downstream) tasks and involves fine-tuning on one of the OOD tasks in the set then testing on the rest. We find that though continual learning methods help, fine-tuning reduces robustness across pretrained models.
Surprisingly, models pretrained on the largest and most diverse datasets (\eg, LAION-2B) exhibit both larger robustness losses and lower absolute robustness after fine-tuning on small datasets, relative to models pretrained on smaller datasets.
We observe this collapse in contrastively pretrained (CLIP) models and their fine-tuned variants, where it grows with pretraining scale; the supervised models we test do not exhibit it. These findings suggest that starting with the strongest foundation model is not necessarily the best approach for performance on specialist tasks\footnote{\url{https://jd730.github.io/projects/ImageNet-RIB}}.

%% file: sections/1_introduction.tex

\vspace{-0.3cm}
\section{Introduction}\label{sec:intro}
\vspace{-0.3cm}
Deep learning has moved toward training large models with  deeper architectures~\citep{dosovitskiy2021image,he2016deep,jiang2023mistral} on massive datasets~\citep{lin2014microsoft,russakovsky2015image,schuhmann2022laion}. These models exhibit impressive performance and generalization abilities; as a result, it has become common to leverage these models as a foundation for fine-tuning on specific downstream datasets to achieve better performance than training from scratch. Fine-tuning can be done with modest amounts of data, and thus is an attractive approach in applications where not enough data is available.

While this approach capitalizes on the extensive knowledge embedded in pretrained models, it can result in significant loss of that knowledge from catastrophic forgetting~\citep{french1999catastrophic,robins1995catastrophic}. Methods to mitigate this problem involve training only a part of the pretrained model, by linear probing, low-rank adaptation~\citep{hu2021lora}, and visual prompting~\citep{bahng2022exploring}.
However, these methods typically underperform on downstream tasks compared to fine-tuning the entire model. 

\begin{figure}[t!]
    \centering
    \includegraphics[width=0.69\linewidth]{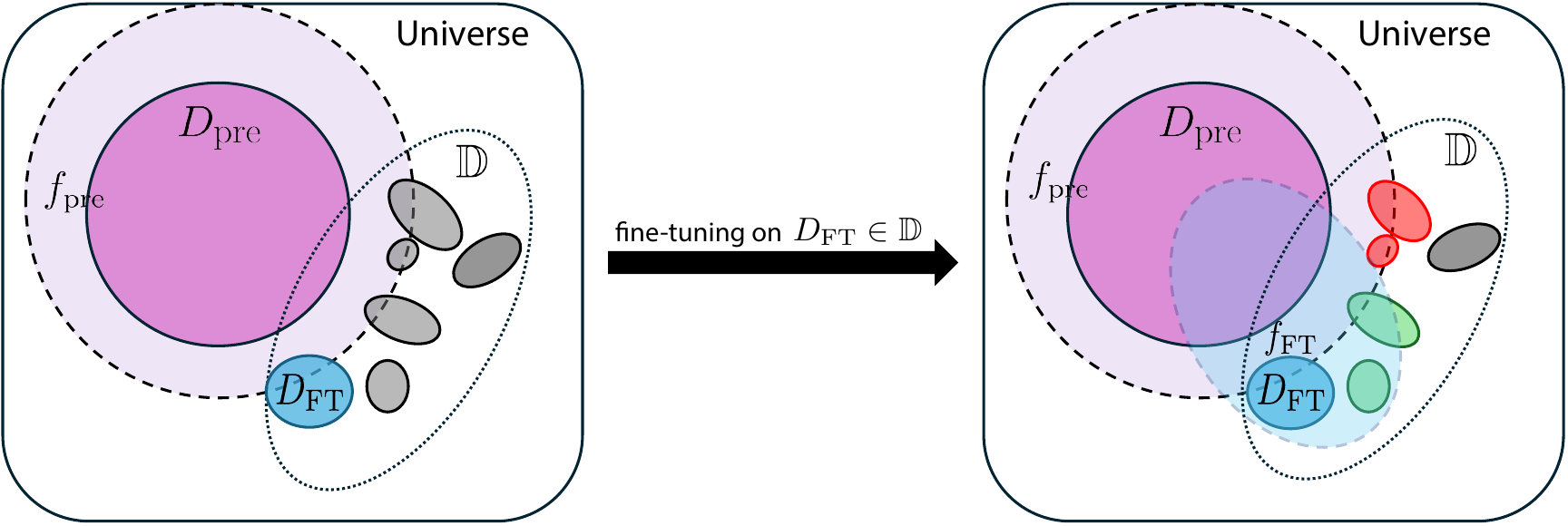}
    \vspace{-0.2cm}
    \caption{
	\textbf{Schematic: How fine-tuning changes the robustness of pretrained models.}
	A model pretrained on the dataset $\Dpre$ (purple solid) has a measure of robustness, generalizing to some out-of-distribution data (purple dashed, $f_\text{pre}$). Dotted gray line:  volume ($\mathbb{D}$) containing a number of related OOD datasets (dark gray ellipsoids). Fine-tuning on one of these datasets ($D_\text{FT}$) shifts $f_\text{pre}$ to $f_\text{FT}$ (blue dashed ellipsoid), increasing performance on $D_\text{FT}$ and some OOD tasks in $\mathbb{D}$ but possibly reducing performance on others (red), thus making the model less robust.
    }
    \label{fig:concept}
\vspace{-0.6cm}
\end{figure}

Fine-tuning also reduces a model's robustness, which we take here to mean the ability to generalize to out-of-distribution (OOD) samples, as the model is optimized for a narrower distribution~(Figure~\ref{fig:concept}).
Model robustness has been extensively studied by using various OOD datasets, typically beginning with an ImageNet pretrained model and evaluating it on OOD datasets that exhibit natural distribution shifts~\citep{taori2020measuring}, such as changes in viewpoints~\citep{barbu2019objectnet}, time~\citep{recht2019imagenet}, styles~\citep{hendrycks2021many,wang2019learning}, or synthetic data based on the original dataset~\citep{hendrycks2019benchmarking,salvador2022imagenetcartoon}.

\begin{wrapfigure}[18]{r}{0.38\textwidth}
    \centering
    \includegraphics[width=0.34\textwidth]{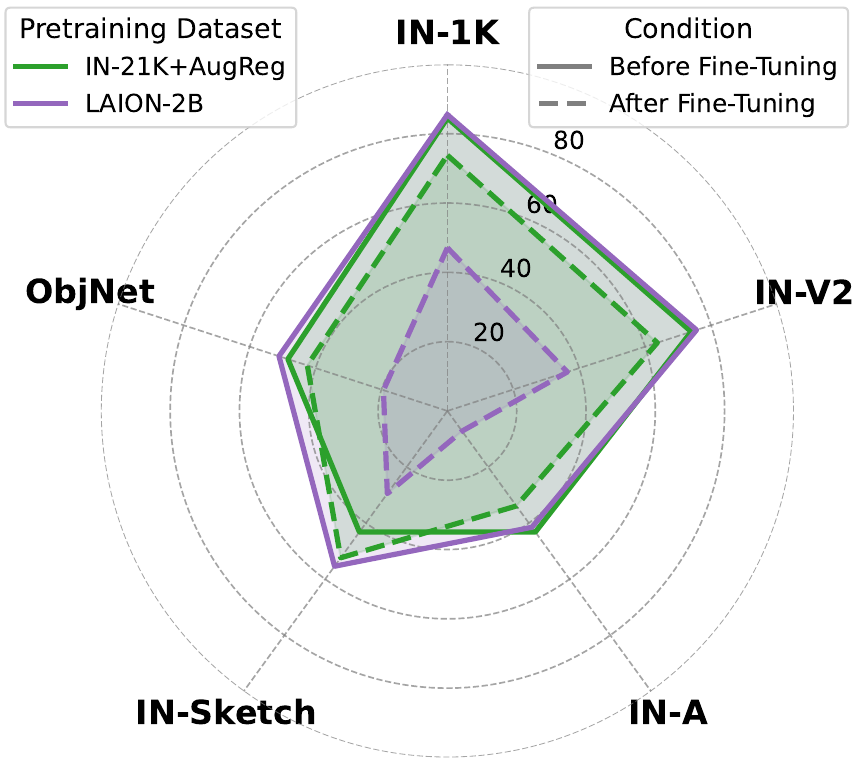}
    \vspace{-0.3cm}
    \caption{{\bf OOD accuracy (robustness) of a ViT-B/16 model pretrained on two different datasets (LAION-2B, IN-21K with AugReg), before and after fine-tuning on ImageNet-R.}}
    \label{fig:rader}
\end{wrapfigure}

We observed that the ViT-B/16 CLIP~\citep{radford2021learning} pretrained on LAION-2B suffers from more severe catastrophic forgetting on OOD datasets after fine-tuning on ImageNet-R~\citep{hendrycks2021many} compared to the same model pretrained on ImageNet-21K~\citep{ridnik2021imagenet} with AugReg~\citep{steiner2022how}, despite their initially similar performance (Figure~\ref{fig:rader}). 
Conversely, the ImageNet-21K pretrained model exhibits improved performance on ImageNet-Sketch~\citep{wang2019learning}.

To analyze why models pretrained on a smaller dataset have better OOD generalizability after fine-tuning, and the effect of the relationship between fine-tuning dataset and OOD datasets, we introduce ImageNet-RIB (Robustness Inheritance Benchmark), a new benchmark designed to assess the robustness of fine-tuned models across diverse downstream and evaluation OOD dataset pairs related to ImageNet.
For each experiment, we fine-tune a pretrained model on a downstream dataset sampled from ImageNet OOD datasets and evaluate its performance on the remaining OOD datasets.
This process is repeated across all available datasets to thoroughly assess how well the model retains robustness after fine-tuning.
We also employ a variety of fine-tuning strategies, including vanilla fine-tuning, linear probing (fine-tuning the last layer only), LoRA~\citep{hu2021lora}, regularization-based continual learning methods~\citep{li2017learning,zenke2017continual}, and robust fine-tuning methods~\citep{kumar2022fine,wortsman2022model,wortsman2022robust}.

Interestingly, pretraining on LAION-2B, despite its size and diversity, does not always yield the best results when fine-tuned on downstream datasets, suggesting that starting with large, rich datasets may not always be the optimal approach for preserving robustness, especially when the downstream dataset size is small.
This problem occurs in the LAION-400M pretrained model, but not in the LAION-100M pretrained model.
Our experimental results also show that the combination of regularization-based continual learning methods with model soup~\citep{wortsman2022model}  achieves the best performance in the benchmark, while linear probing performs the best when using LAION-2B pretrained models.
Furthermore, our findings indicate that continual learning methods not only mitigate catastrophic forgetting related to the pretraining dataset but also enhance robustness when compared to standard fine-tuning.
This improvement is attributed to leveraging the distributional properties of both pretraining and fine-tuning datasets.

In summary, the contributions of this paper are three-fold:
\vspace{-0.3cm}
\begin{itemize}
    \item We show that models pretrained on richer and larger datasets can have worse robustness after fine-tuning than models pretrained on smaller datasets if the fine-tuning dataset size is small.
    This degradation worsens with pretraining scale across CLIP-pretrained and fine-tuned models.
    \item  We propose ImageNet-RIB, a new benchmark leveraging multiple ImageNet-based OOD datasets to quantify the robustness of fine-tuned models in comparison to pretrained models. 
    \item We demonstrate that regularization-based continual learning methods improve robustness by leveraging both the pretraining and fine-tuning dataset distributions. This improvement is amplified when combined with robust fine-tuning methods. 
\end{itemize}

%% file: sections/2_related.tex
\section{Related Work}\label{sec:related}

\paragraph{Robust Fine-Tuning Analysis.}\label{sec:related:robust_finetuning_analysis}
Robust fine-tuning aims to preserve the pretrained model's robustness to out-of-distribution (OOD) datasets, such as variations in viewpoint~\citep{barbu2019objectnet}, style~\citep{hendrycks2021many,wang2019learning}, and temporal distribution shifts~\citep{recht2019imagenet}, during the fine-tuning.
Large-scale transfer-learning studies~\citep{wenzel2022assaying,andreassen2022evolution} fine-tune thousands of ImageNet-pretrained models across many in-distribution (ID),  OOD dataset pairs and show that ID and OOD accuracies typically co-vary, with truly ``effectively robust'' models being rare. \citet{ramanujan2023connection} directly vary properties of the pre-training distribution and, on WILDS~\citep{koh2021wilds}, find that increasing pre-training data quantity improves downstream robustness, whereas additional per-class diversity has a limited effect.
\citet{mai2024fine} show that when fine-tuning covers only a subset of classes, the accuracy drop on the absent classes largely reflects logit miscalibration rather than degraded features, and can be substantially recovered by post-hoc calibration.
\citet{li2025robustness} characterize the accuracy--robustness trade-offs of parameter-efficient fine-tuning methods, considering both adversarial and OOD robustness.
Both studies fix the pretraining distribution.
We ask when robustness acquired during pretraining is inherited, degraded, or inverted by fine-tuning, and we find that models pretrained on the largest web-scale datasets (e.g., LAION-2B) can become less robust after fine-tuning than models pretrained on smaller curated datasets.

\paragraph{Robust Fine-Tuning Benchmark.}\label{sec:related:robustness}

\citet{taori2020measuring} proposed a benchmark that fine-tunes ImageNet-pretrained models on ImageNet-1K and evaluates robustness changes on multiple existing ImageNet-based OOD datasets. Though widely used~\citep{kumar2022fine,wortsman2022model,wortsman2022robust}, this setting fixes fine-tuning datasets and ignores pretraining dataset difference, which limits the analysis of how different downstream datasets or curricula affect robustness.
\citet{shi2023effective} extend this to joint training on two datasets (ImageNet-1K with CIFAR-10~\citep{krizhevsky2009learning} or YFCC~\citep{thomee2016yfcc100m}), but still with a single pretraining distribution.
Our benchmark is complementary to these studies. Rather than fixing the pretraining distribution and focusing on new algorithms, we vary the pretraining dataset (from small curated to large web-scale) while holding a family of related downstream datasets and OOD evaluation suites fixed.
Please refer to Appendix~\ref{supp:related} for comparison with single domain adaptation.

\paragraph{Robust Fine-Tuning Method.}
\citet{wortsman2022model} demonstrate that averaging the parameters of multiple trained models improves both in-distribution and OOD performance.
WiSE-FT~\citep{wortsman2022robust} further shows that linearly interpolating the weights of pretrained CLIP and ImageNet-1K fine-tuned CLIP improves robustness, although it requires tuning the interpolation ratio for optimal performance.
LP-FT~\citep{kumar2022fine} fine-tunes the last layer (linear probing) first and then fine-tunes the entire network.
\citet{goyal2023finetune} show that contrastive learning using a text encoder in fine-tuning improves robustness.
In this paper, we first adopt regularization-based continual-learning methods and combine them with existing  method in robust fine-tuning.

\vspace{-0.1cm}
\paragraph{Continual Learning.}\label{supp:related:cont}
Continual learning aims to enable models to learn new tasks without forgetting previously learned knowledge. Existing approaches can be broadly categorized into three types: regularization-based methods, replay-based methods, and architecture-based methods.
Regularization-based methods~\citep{cheung2019superposition,kirkpatrick2017overcoming,li2017learning,zenke2017continual} use additional loss terms to limit changes to the model's parameters, ensuring that previously learned knowledge is retained. 
For instance, EWC~\citep{kirkpatrick2017overcoming} employs the Fisher information matrix to determine the importance of each parameter, helping to preserve critical weights from earlier tasks. 
LwF~\citep{li2017learning} uses knowledge distillation to transfer outputs from a model trained on past tasks to guide learning new tasks.
Replay-based methods~\citep{robins1995catastrophic} mitigate catastrophic forgetting by creating a replay buffer that contains a subset of previous task data or synthetic data~\citep{van2020brain} and a model is trained on the buffer along with a new task.
Techniques such as reservoir sampling, reinforcement learning~\citep{rebuffi2017icarl}, and gradient-based selection~\citep{aljundi2019gradient} help efficiently manage memory and select important data.
Architecture-based methods modify the model’s structure to accommodate new tasks by dynamically growing networks~\citep{rusu2016progressive,wang2022foster,yan2021dynamically}.
In our work, we focus on regularization-based continual learning methods to ensure a fair comparison with other fine-tuning approaches.

%% file: sections/3_method.tex

\begin{figure*}[t!]
    \centering
    \includegraphics[width=0.85\linewidth]{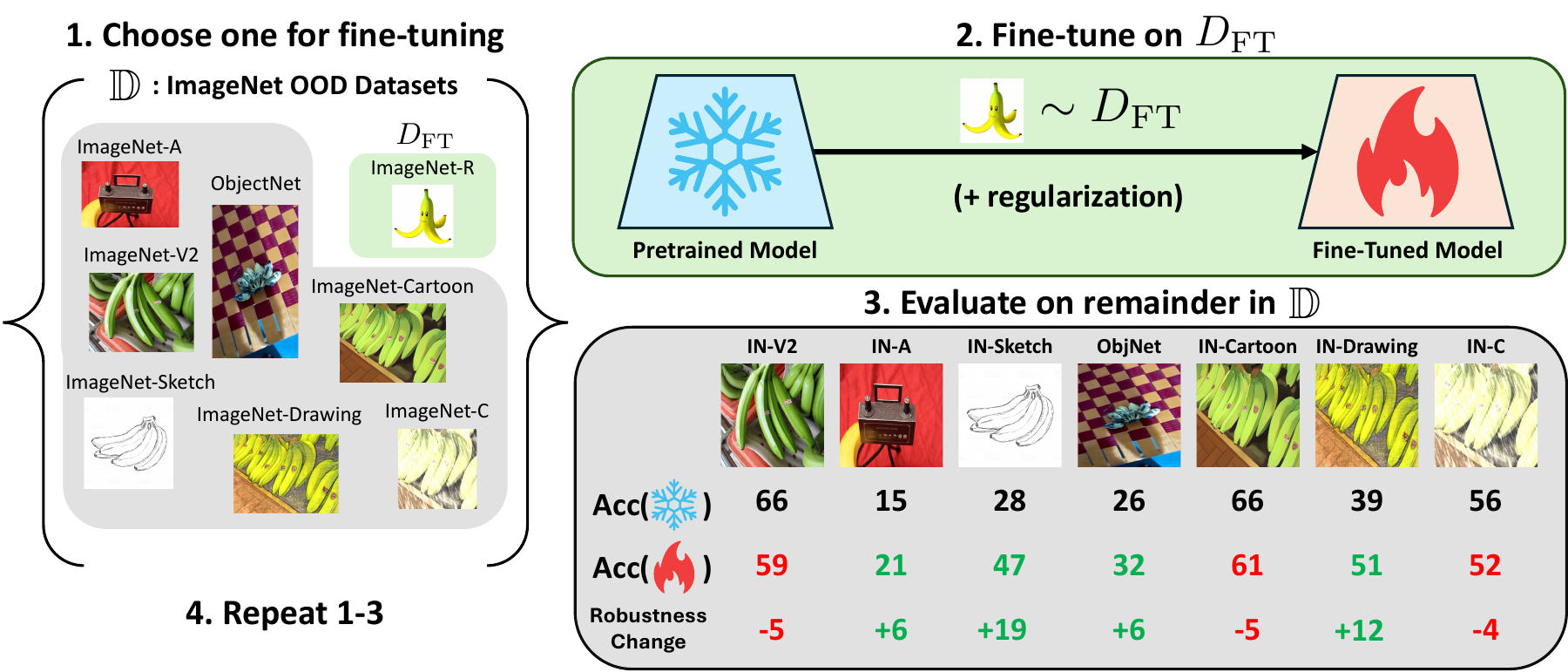}
    \vspace{-0.2cm}
\caption{
\textbf{ImageNet-RIB benchmarking process.}
(1) We define a set $\mathbb{D}$ of ImageNet OOD datasets. We select one for fine-tuning, $D_\text{FT}$, then assess the performance of the pretrained model on $\mathbb{D} \setminus D_\text{FT}$. (2) After fine-tuning the pretrained model on $D_\text{FT}$, we (3) re-assess its performance on $\mathbb{D} \setminus D_\text{FT}$ and compute the robustness change. 
(4) This process is repeated until each dataset in $\mathbb{D}$ has been chosen once as the fine-tuning dataset, ensuring a detailed evaluation of fine-tuning's impact on robustness. 
}
\label{fig:imagenet-rib}
\vspace{-0.4cm}
\end{figure*}

\vspace{-0.2cm}
\section{ImageNet Robustness Inheritance Benchmarking (ImageNet-RIB)} \label{sec:benchmark}
\vspace{-0.2cm}
We propose the ImageNet-RIB (Robustness Inheritance Benchmark), a novel benchmark designed to measure robustness changes using existing ImageNet-related out-of-distribution (OOD) datasets as both fine-tuning and evaluation datasets.
ImageNet-RIB fine-tunes pretrained models on various datasets, then evaluates robustness to other OOD datasets in the benchmark (Figure~\ref{fig:imagenet-rib}), offering a more comprehensive understanding of robustness fine-tuning.

\vspace{-0.1cm}
\subsection{Benchmark Protocol and Robustness Metric} 
\vspace{-0.2cm}

\paragraph{Protocol} 
Figure~\ref{fig:imagenet-rib} illustrates the protocol of our benchmark.
Given a set of out-of-distribution (OOD) datasets $\mathbb{D} = \{D_1, D_2, ..., D_n\}$, we select one to use as a fine-tuning dataset $D_\text{FT}$ for a pretrained model. We evaluate the model's performance on $\mathbb{D} \setminus D_\text{FT}$ before and after fine-tuning on $D_\text{FT}$, and compute the robustness change. This process is repeated by selecting each dataset in $\mathbb{D}$ as the fine-tuning dataset.

\paragraph{Metric}
We define the robustness improvement score ($RI$) as the average relative robustness~\citep{taori2020measuring}.
Specifically, $RI$ measures the accuracy difference between fine-tuned and pretrained models on OOD datasets.
Formally, robustness improvement ($RI$) after fine-tuning on $D_i (=D_\text{FT})$ is defined as:
\begin{equation}
	RI_i = \frac{1}{n-1} \sum_{j=1, j \neq i}^n A^{(j)}_i - A_\text{pre}^{(j)},
\end{equation}
where $A^{(j)}_\text{pre}$ and $A^{(j)}_i$ denote the average accuracies of pretrained and fine-tuned models on $D_j$, respectively.
In addition to relative robustness, effective robustness~\citep{taori2020measuring} is an alternative metric commonly used to evaluate OOD performance~(see Appendix~\ref{supp:effective}).
Effective robustness measures how much the accuracy of a model deviates from an expected baseline, typically using a reference in-distribution dataset (\eg, ImageNet-1K).
We use relative robustness in this benchmark to facilitate direct comparisons between different fine-tuning methods and initial pretraining datasets.
We summarize the overall robustness improvement across all datasets as the mean robustness improvement ($mRI$).

\subsection{Dataset Suites}
We leverage all existing ImageNet OOD datasets to construct $\mathbb{D}$: ImageNet-V2~\citep{recht2019imagenet}, ImageNet-A~\citep{hendrycks2021natural}, ImageNet-Drawing~\citep{salvador2022imagenetcartoon}, ImageNet-Cartoon~\citep{salvador2022imagenetcartoon}, and ImageNet-Sketch~\citep{wang2019learning}, ObjectNet~\citep{barbu2019objectnet}, and ImageNet-C~\citep{hendrycks2019benchmarking}. ObjectNet and ImageNet-C were originally designed solely for evaluating the OOD performance of ImageNet pretrained models, with restrictions on their use for training, however we extend their application in this benchmark by fine-tuning models on these datasets and evaluating their robustness on other OOD datasets.
For detailed descriptions of each dataset, please refer to Appendix~\ref{supp:dataset}.
StanfordCars~\citep{krause20133d} dataset is also used as a showcase of a dataset with different label sets in Appendix~\ref{supp:stanford_cars}.

%% file: sections/4_exp.tex

\section{Experiments}\label{sec:exp}
We use ImageNet-RIB to evaluate how robustly pretrained models fine-tune on downstream datasets, identifying which fine-tuning methods perform best across pretraining sources.

\subsection{Experimental Details}\label{subsec:exp_details}

\paragraph{Pretrained Models.}
We use several architectures of Vision Transformer (ViT)~\citep{dosovitskiy2021image} and ResNet~\citep{he2016deep}.
The models are pretrained on ImageNet-1K~\citep{russakovsky2015image}, or ImageNet-21K~\citep{ridnik2021imagenet} and then fine-tuned on ImageNet-1K.
The standard data augmentation and regularization technique for ViT, AugReg~\citep{steiner2022how} can be used for training on ImageNet-1K or ImageNet-21K.
We also use  ImageNet-1K with Sharpness Aware Minimization (SAM)~\citep{chen2022vision}, ImageNet-21K-P~\citep{ridnik2021imagenet} pretrained models.
Alternatively, some models are pretrained on LAION-2B~\citep{schuhmann2022laion} or OpenAI internal dataset (400 million data)~\citep{radford2021learning}, followed by fine-tuning on ImageNet-1K.
In other words, \textit{all pretrained models are trained on ImageNet-1K before experiments} to directly leverage its classifier unless specified.
For simplicity, we refer to them by the names of the first pretraining datasets (\eg, LAION-2B).
We also evaluate pretrained CLIP models with a zero-shot classifier without fine-tuned on ImageNet-1K in Section~\ref{supp:zeroshot}.
We focus on ImageNet-1K with AugReg pretrained ViT-B/16 and results for other pretrained models are reported in Appendix~\ref{supp:various}.

\paragraph{Fine-tuning Methods.}
We employ standard fine-tuning methods, regularization-based continual learning methods for measuring performance on the proposed benchmark.
The fine-tuning methods we evaluate include vanilla fine-tuning (FT), Linear Probing, LoRA~\citep{hu2021lora}, Visual Prompt~\citep{bahng2022exploring}, LwF~\citep{li2017learning}, and EWC~\citep{kirkpatrick2017overcoming}\footnote{We do not use other continual learning methods as the pretraining dataset is not accessible, and to ensure a fair comparison with other methods.}.
Because we are using ResNets, we do not use LoRA, which was designed for ViT.
We also employ robust fine-tuning methods: LP-FT~\citep{kumar2022fine}, WiSE-FT~\citep{wortsman2022robust}, and uniform model soup~\citep{wortsman2022model}, which averages the parameters of a pretrained model, a vanilla fine-tuned model (FT), LwF, and EWC.
We denote the uniform model soup, MS:PRE-FT-EWC-LwF to reveal the source of parameters.
In Section~\ref{supp:zeroshot}, we also use FLYP~\citep{goyal2023finetune}.

\paragraph{Training.}
Each pretrained model is fine-tuned on $D_\text{FT}$ for 10 epochs with a batch size of 64. We use stochastic gradient descent (SGD) with a learning rate of 0.001 and a momentum of 0.9 with cosine annealing~\citep{loshchilov2017sgdr}.
An ablation over the optimizer and learning rate is provided in Appendix~\ref{supp:optim_recipe}.
Visual Prompt is trained for 10 epochs with a learning rate of 40 without weight decay, following \citet{bahng2022exploring}.
We also evaluate models on ImageNet-RIB with a train-validation split of the fine-tuning dataset and select the best-performing models on the validation set for evaluation in Appendix~\ref{supp:split}.
Please refer to Appendix~\ref{supp:exp_details} and the code repository for detailed implementation.

\begin{table}[t!]
\centering
\caption{
Average robustness of the ViT-B/16 model pretrained on various datasets, assessed on the datasets in ImageNet-RIB set $\mathbb{D}$. LAION-2B pretraining exhibits the highest robustness.
}
 \vspace{-0.2cm}
\label{tab:pre_b16}
\setlength{\tabcolsep}{3pt}
\resizebox{0.8\columnwidth}{!}{
\begin{tabular}{c|c|cccccccc}
\toprule
Pretraining Dataset & ImageNet-1K & IN-V2 &  IN-A & IN-R & IN-Sketch  & ObjNet  & IN-Cartoon & IN-Drawing & IN-C \\ \midrule
IN-1K + AugReg & 79.2 & 66.4 & 15.0 & 38.0 & 28.0 & 33.0 & 66.2 & 39.1 & 56.0 \\
IN-21K&  81.8 & 71.4 & 32.0 & 47.3 & 35.8 & 42.5 & 69.4 & 44.1 & 58.3 \\
IN-21K + AugReg & 84.5 & 74.0 & 43.2 & 56.8 & 43.2 & 48.4 & 75.1 & 54.9 & \textbf{66.5} \\
OpenAI & 85.3 & \textbf{75.7} & \textbf{47.3} & 65.9 & 50.9 & 50.7 & 76.3 & 55.7 & 62.6 \\
LAION-2B& \textbf{85.5} & 75.6 & 41.5 & \textbf{68.8} & \textbf{55.4} & \textbf{51.1} & \textbf{78.2} & \textbf{58.4} & 63.0 \\
\bottomrule
\end{tabular}
}
\vspace{-0.2cm}
\end{table}

\begin{table}[t!]
\centering
\caption{$RI$ and $mRI$ of ViT-B/16 pretrained on ImageNet-1K and AugReg, fine-tuned on each of the datasets in the ImageNet-RIB set  $\mathbb{D}$.}
\label{tab:ri_b16_in1k}
\vspace{-0.2cm}
\resizebox{0.8\columnwidth}{!}{
\begin{tabular}{c|c|cccccccc}
\toprule
\multirow{2}{*}{Method} & \multirow{2}{*}{$mRI$} &\multicolumn{8}{c}{RI on specific $D_\text{FT}$} \\
  &  & IN-V2 &  IN-A & IN-R & IN-Sketch  & ObjNet  & IN-Cartoon & IN-Drawing & IN-C \\ \midrule
FT &1.3 & 2.9 & -4.0 & 2.8 & 4.4 & -2.7 & 0.6 & 0.4 & 5.9 \\
Linear Probing &0.7 & 0.1 & -0.1 & 0.8 & 1.2 & 0.3 & 0.2 & 0.1 & 3.2 \\
Visual Prompt &-4.5 & -2.3 & -9.1 & -4.9 & -1.6 & -11.2 & -3.9 & -4.3 & 1.7 \\
LoRA &0.9 & 0.2 & 0.4 & 1.1 & 2.6 & 0.3 & -0.1 & 1.3 & 1.1 \\
EWC &2.8 & 2.9 & -0.2 & 5.2 & 4.4 & 1.4 & 1.6 & 2.8 & 4.3 \\
LwF &3.1 & 2.8 & -0.0 & 6.2 & 4.6 & 0.7 & 1.9 & 2.1 & 6.5 \\
LP-FT &2.3 & \textbf{3.0} & -0.9 & 5.2 & 4.5 & -0.1 & 1.2 & 0.6 & 4.7 \\
WiSE-FT &3.6 & 2.5 & \textbf{0.7} & 7.5 & 4.5 & 2.1 & 2.3 & 3.0 & 6.5 \\
MS &\textbf{3.9} & 2.7 & 0.7 & \textbf{7.8} & \textbf{5.0} & \textbf{2.2} & \textbf{2.4} & \textbf{3.3} & \textbf{6.7} \\
\bottomrule
\end{tabular}
}
\vspace{-0.2cm}
\end{table}

\subsection{Combination of Continual Learning with Robust Fine-Tuning Methods Perform Best}\label{subsec:robustness_improvement}
\paragraph{Baseline}
We start with the baseline of assessing model performance on the set of OOD datasets without any fine-tuning.
Models pretrained on larger and more diverse datasets have better performance on both ImageNet-1K and downstream datasets as shown in Table~\ref{tab:pre_b16}.
However, the ImageNet-21K with AugReg pretrained model achieves better performance on ImageNet-C than LAION-2B pretrained model since AugReg includes several corruptions in ImageNet-C (\eg, brightness and  contrast).

\paragraph{Accuracy on OOD Datasets.}
Table~\ref{tab:main_base_16_in1k} presents the accuracy of an ImageNet-1K with AugReg pretrained ViT-B/16 model on OOD datasets before and after fine-tuning with each method on the fine-tuning dataset~(see Table~\ref{tab:inc_base_16_in1k} for individual ImageNet-C corruption).
Continual learning methods and robust fine-tuning methods generally improve performance on most OOD datasets after fine-tuning on the downstream datasets.
Linear probing (LP) exhibits similar increase and decrease patterns as vanilla fine-tuning (FT), with less magnitude as the backbone network is fixed.
Visual Prompt reduces performance even on ImageNet-1K after fine-tuning on synthetic datasets of the ImageNet validation set.
This contrasts with \citet{bahng2022exploring}, which showed robustness to OOD datasets.
This discrepancy stems from differing setups that \citet{bahng2022exploring} tune prompts on frozen CLIP with a zero-shot text classifier.
Visual Prompt is likewise among the most robust methods in our benchmark (Table~\ref{tab:zeroshot}).
In contrast, with the fixed ImageNet-1K linear head used here, this single input-space perturbation fails to transfer and degrades accuracy, even on clean ImageNet-1K.
A strong correlation exists between ImageNet-R, ImageNet-Sketch, and ImageNet-Drawing, as they share drawing and sketch renditions, and ImageNet-R and ImageNet-Sketch share images.
Fine-tuning on ImageNet-C improves performance on other synthetic datasets, but not vice versa, due to its diverse corruptions and severities.

\paragraph{Robustness Improvement.}
Individual robustness improvement scores ($RI$) after fine-tuning on each OOD dataset with ImageNet-1K with AugReg pretrained ViT-B/16 also show that MS:PRE-FT-EWC-LwF consistently performs the highest in most datasets, followed by WiSE-FT as demonstrated in Table~\ref{tab:ri_b16_in1k}.
This is because they directly use the weights of pretrained models, thus taking advantage of their robustness.

\paragraph{Mean Robustness Improvement.}
The combination of continual learning methods with weight averaging (MS:PRE-FT-EWC-LwF) achieves the highest or second-highest mean robustness improvement ($mRI$) across different backbones pretrained on ImageNet-based datasets as shown in Table~\ref{tab:mri}.
Moreover, end-to-end continual learning methods show comparable performance to the multi-stage method~\citep{kumar2022fine} or the post-hoc robustness method~\citep{wortsman2022robust}.
This shows the potential of continual learning methods in the field of robust fine-tuning.
The robustness of linear probing and Visual Prompt remains relatively unchanged since they do not modify the models' weights significantly but their performance on the downstream dataset tends to be worse (see Appendix~\ref{supp:downstream}).
Consequently, they have much better performance with LAION-2B pretrained models compared to other methods, which show a significant robustness decrease.

\begin{figure}[t!]
    \centering
    \includegraphics[width=0.5\linewidth]{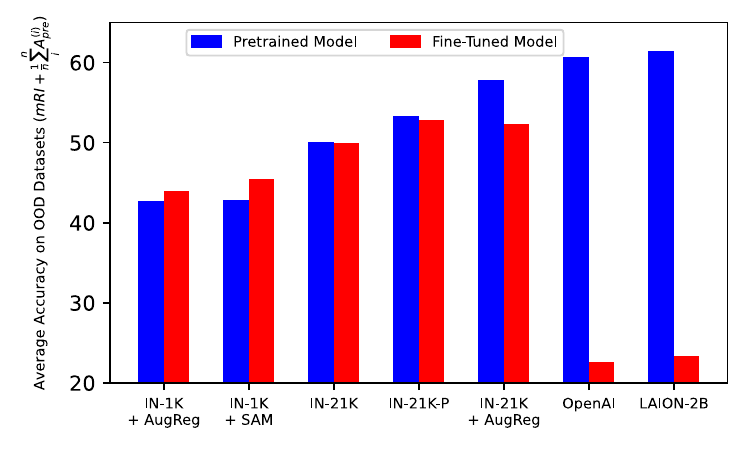}
    \vspace{-0.3cm}
\caption{
\textbf{Severe robustness loss from fine-tuning models pretrained on LAION-2B and OpenAI relative to fine-tuning models pretrained only smaller datasets.} 
The average accuracy on OOD datasets before (blue) and after (red) vanilla fine-tuning (see Figure~\ref{fig:mri} for other methods).
The red bars are calculated by evaluating fine-tuned pretrained models on $\mathbb{D}$. The blue bars are the pretrained models' mean accuracy on $\mathbb{D}$. 
Fine-tuning models pretrained on LAION-2B and OpenAI causes severe robustness loss, leading to worse absolute performance on $\mathbb{D}$  than after fine-tuning an AugReg model pretrained on ImageNet-1K.
Conversely, a model pretrained on ImageNet-1K with AugReg actually exhibits improved robustness after fine-tuning. Note that the difference between red and blue bars is $mRI$.
}
    \vspace{-0.2cm}
\label{fig:mri_ft}
\end{figure}

\input{method_first_tables/mri}

\subsection{Paradoxically, Models pretrained on the Largest Datasets  Do Worst After Fine-Tuning}\label{subsec:paradox}
The extent of robustness degradation increases with the size and diversity of the pretraining dataset, as illustrated in Table~\ref{tab:mri} and Figure~\ref{fig:mri_ft}.
As a result, fine-tuned models pretrained on larger datasets (\eg, LAION-2B, OpenAI) exhibit worse robustness than those pretrained on smaller datasets when using vanilla fine-tuning.
Similarly, the LAION-400M CLIP model has better robustness than the LAION-2B model with a zero-shot classifier from a text encoder after fine-tuning (see Table~\ref{tab:zeroshot_vit}).
The discrepancy between the LAION-2B and ImageNet-1K + AugReg pretrained models increases when selecting the best-performing epoch for each case instead of using a fixed 10-epoch duration (see Appendix~\ref{supp:split}).
The magnitude of this degradation, however, depends on the fine-tuning setting.
Under AdamW with a learning rate of 0.00001, the collapse is much milder ($mRI$ of the LAION-2B pretrained ViT-B/16: $-38.1 \to -4.8$).
Nevertheless, LAION-2B and OpenAI remain the two least robust pretraining datasets for ViT-B/16 (Appendix~\ref{supp:optim_recipe}).

One possible explanation is that models pretrained on larger, more diverse datasets have more room for performance degradation from catastrophic forgetting as they demonstrate higher initial robustness  (see Table~\ref{tab:pre_b16}).
However, this does not fully explain the pronounced robustness loss observed in OpenAI or LAION-2B pretrained models, particularly when compared to ImageNet-21K with AugReg pretrained models, which exhibit similar initial robustness.
Notably, ImageNet-21K and its variants begin to exhibit robustness degradation, especially when using vanilla fine-tuning.
This could be an early indicator of performance decay in larger pretrained models.
Although ImageNet-21K is the second-largest dataset with 14 million images, it is much smaller than LAION-2B, which contains two billion images.
We hypothesize that this discrepancy in pretraining dataset size contributes to the difference in robustness degradation.

\begin{table}[t!]
\centering
\caption{$mRI$ of ViT-B/16, ViT-B/32, ConvNeXt-Base, and ResNet-50 CLIP models, and SigLip-B/16, SigLip2-B/16 pretrained on various datasets using zero-shot classifier~\citep{radford2021learning}.
}
\vspace{-0.1cm}
\label{tab:zeroshot}
\resizebox{\textwidth}{!}{
\begin{tabular}{c|ccc|cccc|c|ccc|c|c}
\toprule
\multirow{2}{*}{Method}  & \multicolumn{3}{c|}{ViT-B/16}&  \multicolumn{4}{c|}{ViT-B/32} & ConvNeXt & \multicolumn{3}{c|}{ResNet-50} & SigLip & SigLip2 \\ 
 &  {\shortstack[c]{LAION\\400M}}& OpenAI& {\shortstack[c]{LAION\\2B}} & {\shortstack[c]{LAION\\100M}}& {\shortstack[c]{LAION\\400M}} & OpenAI &{\shortstack[c]{LAION\\2B}}  & LAION-2B & {\shortstack[c]{CC\\12M}} & {\shortstack[c]{YFCC\\15M}}& OpenAI &WebLI & WebLI\\
\midrule
FT    &  -32.5 &-31.2 & -37.4 & -7.3 & -23.8 &  -27.0 & -33.0   &-16.9 & -7.1 & -2.6 & -24.1 & -21.3 &-25.7\\
FYLP & -26.0 & -30.9 & -36.6 & -4.6 & -24.5 & -26.9 & -32.8& -17.5 &\textbf{-4.4} & \textbf{-0.7} & -26.1 & -24.0 & -28.7 \\
Visual Prompt & \textbf{-7.6} & -11.0 &  \textbf{-10.2}  & -11.6 & -9.6 & -7.7 & \textbf{-10.4}  &-53.5& -8.9 & -6.3 & \textbf{-10.9} & -26.7 &-25.6\\
LoRA & -47.8 &-49.2 & -52.5 & -35.2 & -41.3& -41.1 &-19.4& - & - & -& - & -47.4 & -20.7\\
EWC  & -8.1& \textbf{-8.8}  & -13.1 & -1.5 & \textbf{-5.4}& \textbf{-7.3} & -10.5    & -5.5 & -8.8 & -5.8 & -17.2 & \textbf{-1.9}& -3.1\\
LwF  & -29.6  & -24.9 & -30.9 & -2.1 & -16.8  & -22.5 & -29.1 & -11.8 & -6.2 & -1.7 & -22.1 & -12.9& -17.2 \\
WiSE-FT & -19.5 &-14.4& -23.2& 1.7 & -9.5& -11.9& -17.5  &-4.3& -8.3 & -3.4 & -29.1 & -4.3 & -7.3\\
MS & -20.4& -11.3   & -18.9 & \textbf{2.3} & -15.0 & -9.4 & -15.0  &\textbf{-3.9}& -10.3 & -3.5 & -35.1 & -2.2 & \textbf{-4.2}\\
\bottomrule
\end{tabular}
}
\vspace{-0.1cm}
\end{table}

\begin{table}[t!]
\centering
\caption{
Average zero-shot accuracy on ImageNet-1K validation set and each OOD dataset using pretrained ViT-B/16 and ResNet-50 CLIP models.
All pretrained weights are acquired from \textsf{open\_clip} library.
 }
 \vspace{-0.1cm}
\label{tab:zeroshot_pre}
\resizebox{0.9\columnwidth}{!}{
\begin{tabular}{c|c|c|c|cccccccc}
\toprule
Architecture& Pretraining Dataset & IN-1K & Avg. OOD & IN-V2 &  IN-A & IN-R & IN-Sketch  & ObjNet  & IN-Cartoon & IN-Drawing & IN-C \\ \midrule
\multirow{3}{*}{ViT-B/16} &LAION-400M &67.1 &50.2 &59.6 &33.1 &77.9 &52.4 &46.1 &56.1 &36.4 &40.2\\
&OpenAI &64.4 &48.8 &57.8 &44.4 &73.5 &44.3 &50.5 &48.2 &33.3 &38.2\\
&LAION-2B &70.2 &53.5 &62.3 &38.0 &80.6 &56.1 & 50.8 &59.2 &37.5 &43.0\\
\midrule
\multirow{4}{*}{ViT-B/32} &LAION-100M &52.5 & 38.6 & 44.5 & 14.6 & 64.5 & 39.8 & 30.0 & 43.9 & 44.5 &27.3\\
&LAION-400M &60.2 & 42.6 & 52.4 & 19.6 & 70.8 & 46.4 & 38.9 &48.4 & 29.4  & 35.1\\
&OpenAI & 59.6 & 41.4 & 52.9 & 28.3 & 67.1 & 40.4 & 31.6 & 46.1 & 29.1 & 35.9\\
&LAION-2B & 66.6 & 48.7 & 58.1 & 26.3 & 76.4 & 53.7 & 44.7 & 55.8 & 33.2 & 41.6\\
\midrule
ConvNeXT & LAION-2B & 70.8 & 57.2 & 62.4 & 40.1 & 80.7 & 57.1 & 52.2 & 58.5 & 62.4 & 44.5\\
\midrule
\multirow{3}{*}{ResNet-50} &CC-12M & 35.9 &21.4 & 30.6 & 7.5 & 44.6 & 23.5 & 21.8 & 23.8 & 9.1 & 10.1\\
& YFCC-15M & 32.3 & 15.3 &28.0 & 13.7 & 22.2 & 7.3 & 16.7 & 17.4 & 6.8 & 10.0\\
& OpenAI &57.9 & 36.8 & 50.9 & 23.3 & 60.1 & 34.6 & 36.9 & 40.2 & 22.4 & 25.6\\
\midrule
SigLIP & WebLI & 76.1 & 60.6 & 69.0 & 45.0 & 90.2 & 67.9 & 50.8 & 67.4 & 47.5 & 47.1 \\
\midrule
SigLip2 & WebLI & 69.7 & 59.3 & 63.4 & 54.1 & 83.9 & 61.8 & 53.8 & 63.3 & 47.0 & 47.2 \\
\bottomrule
\end{tabular}
}
\end{table}

\subsection{CLIP with Zero-Shot Classifier on ImageNet-RIB}\label{supp:zeroshot}

In Section~\ref{subsec:robustness_improvement}, we evaluated models pretrained on various datasets and subsequently fine-tuned on ImageNet-1K for the ImageNet-RIB benchmark.
Here, we extend this analysis to measure the $mRI$ of CLIP models that bypass ImageNet-1K fine-tuning and are directly fine-tuned on downstream datasets.
We utilize pretrained weights from the \textsf{open\_clip} library~\citep{ilharco_gabriel_2021_5143773} and adopt a zero-shot classifier, as proposed by \citet{radford2021learning}, instead of a linear readout layer.
This choice is driven by the fact that each OOD dataset contains a distinct subset of ImageNet-1K labels, making a unified linear probe impractical.
For example, fine-tuning on ImageNet-A involves only 200 classes, whereas ImageNet-Sketch covers 1,000 classes. Consequently, methods such as Linear Probing and LP-FT are excluded, as zero-shot classifiers do not require additional training.

Table~\ref{tab:zeroshot} reports the $mRI$ of various ViT-B/16, ViT-B/32, ConvNeXt-Base~\citep{liu2022convnet} and ResNet-50 CLIP models and SigLip-B/16~\citep{zhai2023sigmoid}, SigLip2-B/16~\citep{tschannen2025siglip} pretrained on different datasets without fine-tuned on ImageNet-1K.
FLYP~\citep{goyal2023finetune} is effective on ResNet-50 CLIP pretrained on a smaller dataset, while it does not solve a problem in other models.
Consistent with the results in Table~\ref{tab:mri}, we observe that most fine-tuning approaches, including vanilla fine-tuning, significantly degrade robustness when using models pretrained on the large-scale dataset.
However, ViT-B/32 CLIP model pretrained on LAION-100M~\citep{lin2024parrot} and ResNet-50 CLIP models pretrained on Conceptual-12M (CC-12M)~\citep{changpinyo2021cc12m} and YFCC-15M~\citep{thomee2016yfcc100m} do not exhibit this robustness degradation.
This suggests that a CLIP model pretrained on comparatively smaller datasets experiences less catastrophic forgetting in out-of-distribution generalization than a model pretrained on larger datasets. 
Additionally, Table~\ref{tab:zeroshot_pre} presents the zero-shot accuracy of pretrained CLIP models on the ImageNet-1K validation set and each OOD dataset.
Taking into account both the average OOD accuracy and $mRI$, LAION-400M outperforms LAION-2B, achieving a 2.9-point higher OOD accuracy after vanilla fine-tuning (19.0 vs.\ 16.1).

\begin{table}[t!]
\centering
\makebox[0pt][c]{\parbox{\textwidth}{%
    \begin{minipage}[b]{0.48\hsize}\centering
    \caption{
    Average Accuracy on OOD Datasets of each ViT-B/16 CLIP model with zero-shot classifier before and after vanilla fine-tuning (FT).
    LAION-400M pretrained model outperforms LAION-2B pretrained one after fine-tuning.
     }
       \vspace{-0.2cm}
    \label{tab:zeroshot_vit}
    \resizebox{\columnwidth}{!}{
    \begin{tabular}{c|ccc}
    \toprule
    ViT-B/16 CLIP &	OpenAI	&LAION-400M &	LAION-2B\\
    \midrule
    Before Fine-Tuned &	48.8&	50.2&	\textbf{53.5} \\
    After Fine-Tuned &16.3 &	\textbf{19.0}	&16.1 \\
    \bottomrule
    \end{tabular}
    }
 \end{minipage}
    \hfill
    \begin{minipage}[b]{0.48\hsize}\centering
    \caption{
$mRI$ of ViT-B/32 CLIP with zero-shot classifier after vanilla fine-tuning (FT).
Only OpenAI pretrained model has huge negative $mRI$.
 }
\label{tab:zeroshot_res}
\resizebox{\columnwidth}{!}{
\begin{tabular}{c|ccc}
\toprule
ViT-B/32 & LAION-100M & LAION-400M & LAION-2B\\
\midrule
$mRI$ &	\textbf{-7.3} & -23.8 & -33.0 \\
\bottomrule
\end{tabular}
}
    \end{minipage}
}}
\vspace{-0.2cm}
\end{table}

\begin{figure}
  \centering
  \includegraphics[width=0.6\linewidth]{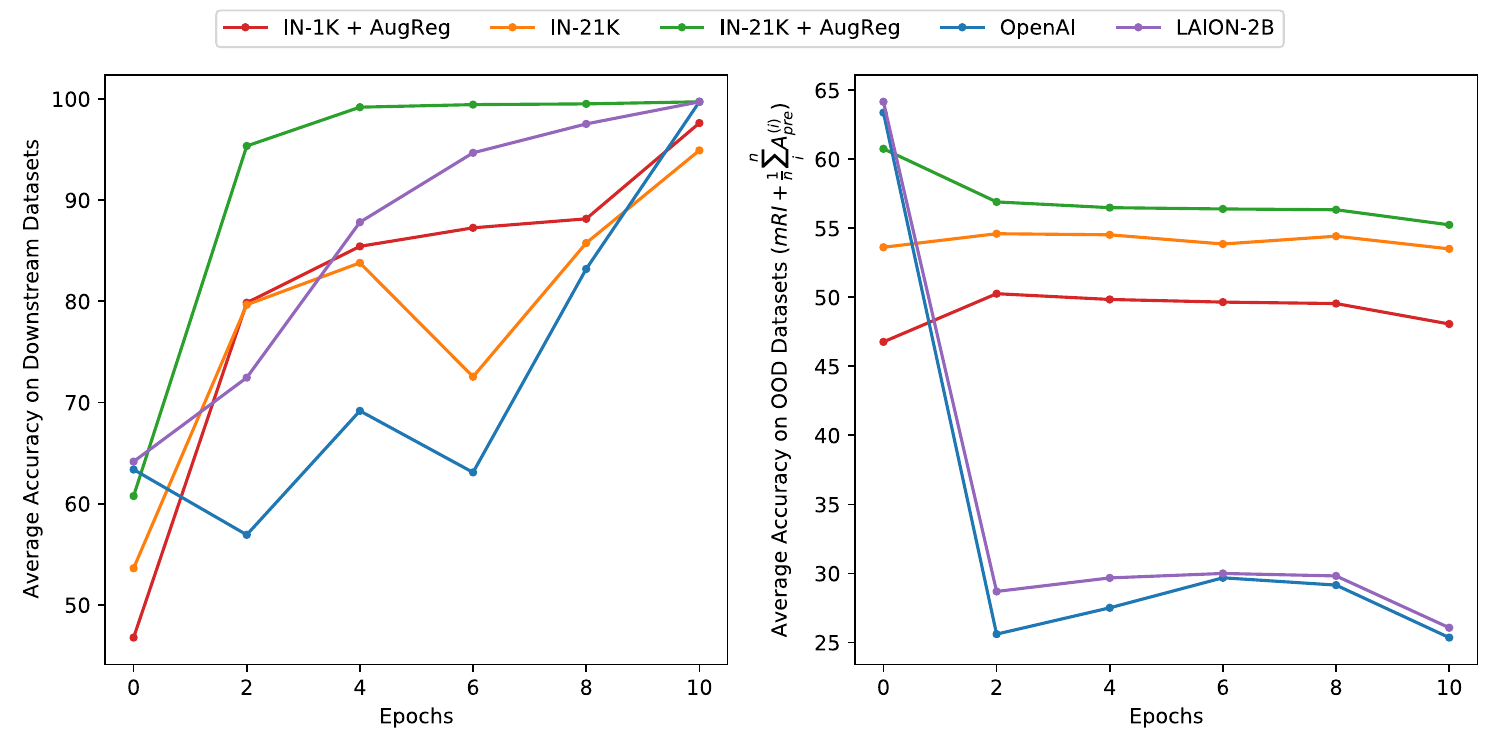}
  \vspace{-0.3cm}
     \caption{
     \textbf{Fine-tuning LAION-2B and OpenAI pretrained models causes severe robustness loss while learning slower than ImageNet-21K pretrained AugReg model.}
     The average accuracy on fine-tuning datasets (left) and the average accuracy on OOD datasets (right) while fine-tuning on the downstream dataset using vanilla fine-tuning method (FT) with ViT-B/16.
     Although these models learn slower than other methods, they suffer from a huge robustness drop even in the early period of fine-tuning.
     }
     \label{fig:progress}
     \vspace{-0.3cm}
\end{figure}%

\subsection{Analysis of Severe Catastrophic Forgetting in Models Pretrained on Large Datasets}\label{subsec:analysis}

\paragraph{CLIP pretrained on Small Dataset Does Not Exhibit Severe Robustness Degradation.}

To discern whether the robustness degradation arises from the CLIP pretraining method itself or from the scale of the pretraining data, we evaluate CLIP models trained on smaller datasets.
The ViT-B/32 CLIP model pretrained on LAION-100M shows much less degradation, whereas counterparts trained on LAION-400M and LAION-2B do (see Table \ref{tab:zeroshot_res}) although they have similar performance before fine-tuning as shown in Table~\ref{tab:zeroshot_pre}.
Similarly, ResNet-50 CLIP models pretrained on CC-12M \citep{changpinyo2021cc12m} and YFCC-15M \citep{thomee2016yfcc100m} remain robust, while the model pretrained on OpenAI’s internal 400M-image dataset exhibits pronounced degradation as we mentioned in Section \ref{supp:zeroshot}.

\paragraph{Overfitting Does Not Drive Robustness Collapse.}\label{supp:progress}
A plausible explanation for the observed decline in robustness during fine-tuning could be early overfitting in LAION-2B and OpenAI pretrained ViT-B/16 models.
We investigate this by tracking robustness performance and average accuracy on the downstream datasets throughout standard fine-tuning (FT).
Figure~\ref{fig:progress} reveals that the ImageNet-21K model pretrained with AugReg learns the fine-tuning dataset faster than other methods, while OpenAI pretrained model shows the slowest learning progression.
Despite this, only the LAION-2B and OpenAI models experience significant OOD robustness degradation.
This finding indicates that overfitting is not the primary driver of catastrophic forgetting in these models.
Moreover, Appendix~\ref{supp:lr} shows that catastrophic forgetting occurs even with a much smaller learning rate.

\begin{figure}
  \centering
  \includegraphics[width=0.5\linewidth]{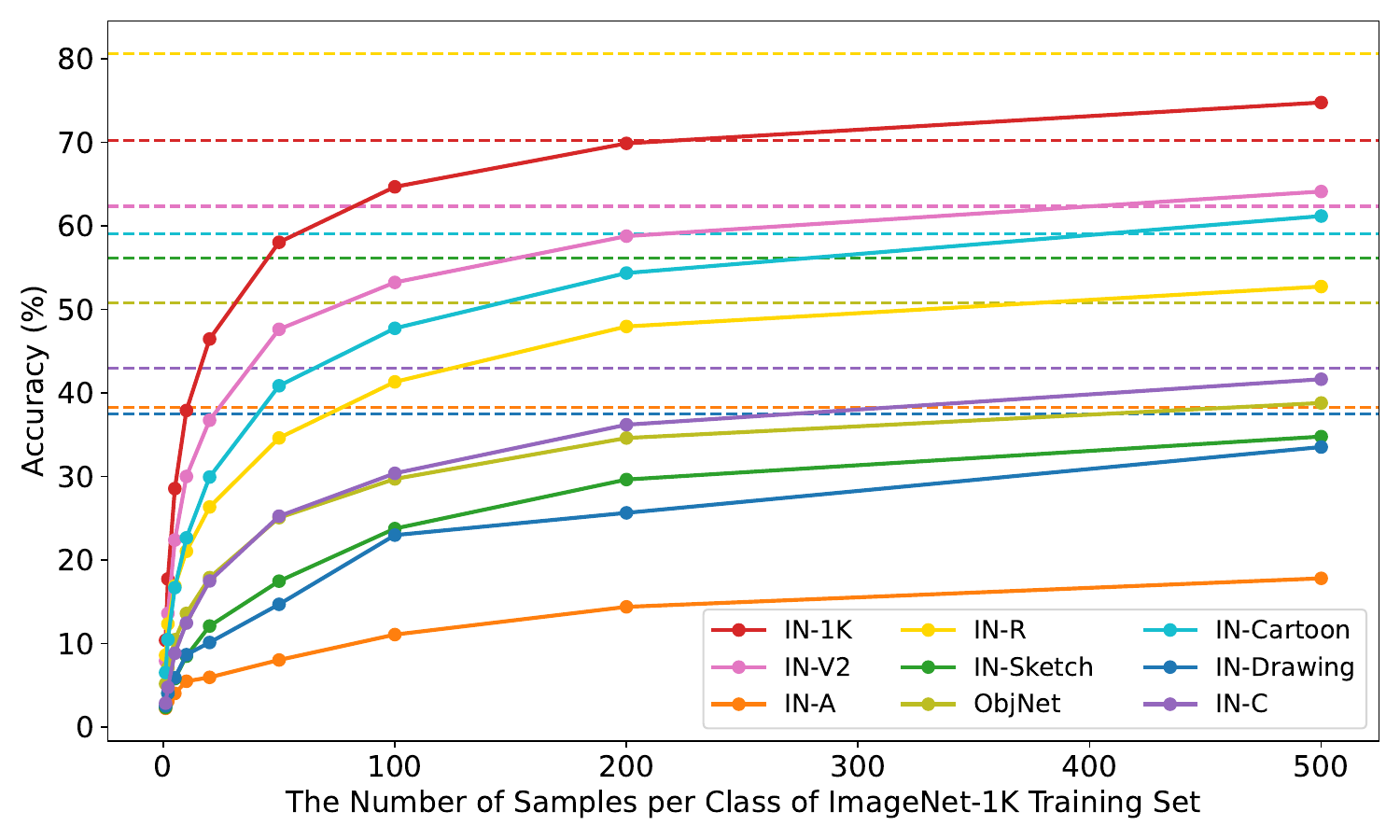}
    \caption{
    \textbf{Fine-tuning on small dataset leads to severe accuracy degradation both in- and out-of distribution.}  
Accuracy on ImageNet-1K validation set and OOD datasets after fine-tuning LAION-2B pretrained ViT-B/16 CLIP model with zero-shot classifier on a small number of images per class of ImageNet-1K training set.
    The dashed line denotes the accuracy of the pretrained model on each dataset.
    }
    \label{fig:ft_1k}
\end{figure}

\paragraph{Fine-Tuning Dataset Texture Does Not Account for Forgetting.}\label{supp:hypothesis_style}
Unlike traditional benchmarks~\citep{taori2020measuring}, which use natural images (\eg, ImageNet-1K), the ImageNet-RIB benchmark incorporates a variety of styles, including cartoons, drawings, and sketches. One may hypothesize that models pretrained on large datasets are susceptible to robustness degradation when fine-tuned on downstream datasets featuring stylized or non-natural images.
However, our findings challenge this hypothesis;  fine-tuning on the ImageNet-1K validation set also leads to similar robustness collapse, even though \textit{all} models are pretrained and then fine-tuned on ImageNet-1K training set (see Table~\ref{tab:ft_in_val}).

\begin{table}[t!]
\centering
\caption{
Accuracy on each dataset of ViT-B/16 pretrained models after fine-tuning on ImageNet-1K validation set.
The parenthesis denotes the difference with pretrained models.
}
\vspace{-0.2cm}
\label{tab:ft_in_val}
\resizebox{0.95\textwidth}{!}{
\begin{tabular}{c|c|ccccc|ccc}
\toprule
Pretraining Dataset & IN-1K & IN-V2 & IN-A & IN-R & IN-Sketch & ObjNet & IN-Cartoon & IN-Drawing & IN-C \\
\midrule
IN-1K + AugReg & 97.5 ({\color{red}+18.3}) & 66.9 ({\color{red}+0.4}) & 23.3 ({\color{red}+8.3}) & 40.9 ({\color{red}+2.9}) & 29.5 ({\color{red}+1.5}) & 37.2 ({\color{red}+4.2}) & 71.1 ({\color{red}+4.9}) & 41.0 ({\color{red}+1.9}) & 59.5 ({\color{red}+3.5})\\
IN-1K + SAM & 87.3 ({\color{red}+7.1}) & 69.4 ({\color{red}+1.2}) & 17.7 ({\color{red}+8.7}) & 41.8 ({\color{red}+1.7}) & 30.1 ({\color{red}+2.4}) & 38 ({\color{red}+3.8}) & 72.1 ({\color{red}+5.2}) & 42.9 ({\color{red}+0.6}) & 56.9 ({\color{red}+2.3})\\
IN-21K & 94.7 ({\color{red}+12.9}) & 71.6 ({\color{red}+0.2}) & 38.5 ({\color{red}+6.5}) & 49.9 ({\color{red}+2.6}) & 36.7 ({\color{red}+0.9}) & 45.2 ({\color{red}+2.7}) & 73.9 ({\color{red}+4.5}) & 44.1 (0.0) & 59.8 ({\color{red}+1.5})\\
IN-21K-P & 96.9 ({\color{red}+12.6}) & 73.0 ({\color{blue}-1.0}) & 41.4 ({\color{red}+7.3}) & 51.5 (0.0) & 39.8 ({\color{blue}-0.4}) & 45.8 ({\color{blue}-0.9}) & 76.4 ({\color{red}+2.9}) & 44.3 ({\color{blue}-0.8}) & 61.7 ({\color{red}+0.3})\\
IN-21K + AugReg & 99.9 ({\color{red}+15.4}) & 70.6 ({\color{blue}-3.4}) & 42.2 ({\color{blue}-1.0}) & 54.1 ({\color{blue}-2.7}) & 39.4 ({\color{blue}-3.8}) & 47.9 ({\color{blue}-0.5}) & 84.5 ({\color{red}+9.4}) & 55.5 ({\color{red}+0.6}) & 69.7 ({\color{red}+3.2})\\
OpenAI & 99.9 ({\color{red}+14.6}) & 59.9 ({\color{blue}-15.8}) & 13.9 ({\color{blue}-33.4}) & 34.9 ({\color{blue}-31.0}) & 19.7 ({\color{blue}-31.2}) & 30.5 ({\color{blue}-20.2}) & 75.0 ({\color{blue}-1.3}) & 33.4 ({\color{blue}-22.3}) & 45.7 ({\color{blue}-16.9})\\
LAION-2B & 99.9 ({\color{red}+14.4}) & 59.4 ({\color{blue}-16.2}) & 12.6 ({\color{blue}-28.9}) & 36.3 ({\color{blue}-32.5}) & 23.4 ({\color{blue}-32.0}) & 30.4 ({\color{blue}-20.7}) & 73.0 ({\color{blue}-5.2}) & 30.6 ({\color{blue}-27.8}) & 41.8 ({\color{blue}-21.2})\\
\bottomrule
\end{tabular}
}
\end{table}

\paragraph{Fine-Tuning Dataset Size is a Major Determinant.}\label{supp:hypothesis_size}

The consistent robustness degradation seen in OpenAI and LAION-2B pretrained models fine-tuned on the ImageNet-1K validation set leads us to hypothesize that the size of the downstream dataset plays a significant role in catastrophic forgetting.
While \citet{ramanujan2023connection} and \citet{fang2022data} demonstrate that CLIP's robustness is primarily attributed to the pretraining dataset size and distribution—rather than contrastive learning—the impact of downstream dataset size remains underexplored.
To investigate, we fine-tune a LAION-2B pretrained CLIP model (not previously fine-tuned on ImageNet-1K) on subsets of the ImageNet-1K training set, using a zero-shot classifier similar to Section~\ref{supp:zeroshot}.
As shown in Figure~\ref{fig:ft_1k}, both ImageNet-1K validation accuracy and OOD performance degrade significantly when fine-tuned on smaller subsets.
This degradation persists across hyperparameter variations, including learning rate and training epochs.
These findings indicate that CLIP models require sufficiently large fine-tuning datasets to maintain robustness against distribution shifts.
Insufficient data in fine-tuning likely exacerbates catastrophic forgetting, highlighting dataset size as a critical factor in mitigating OOD performance decline.

\begin{figure*}[t!]
    \centering
    \includegraphics[width=0.9\linewidth]{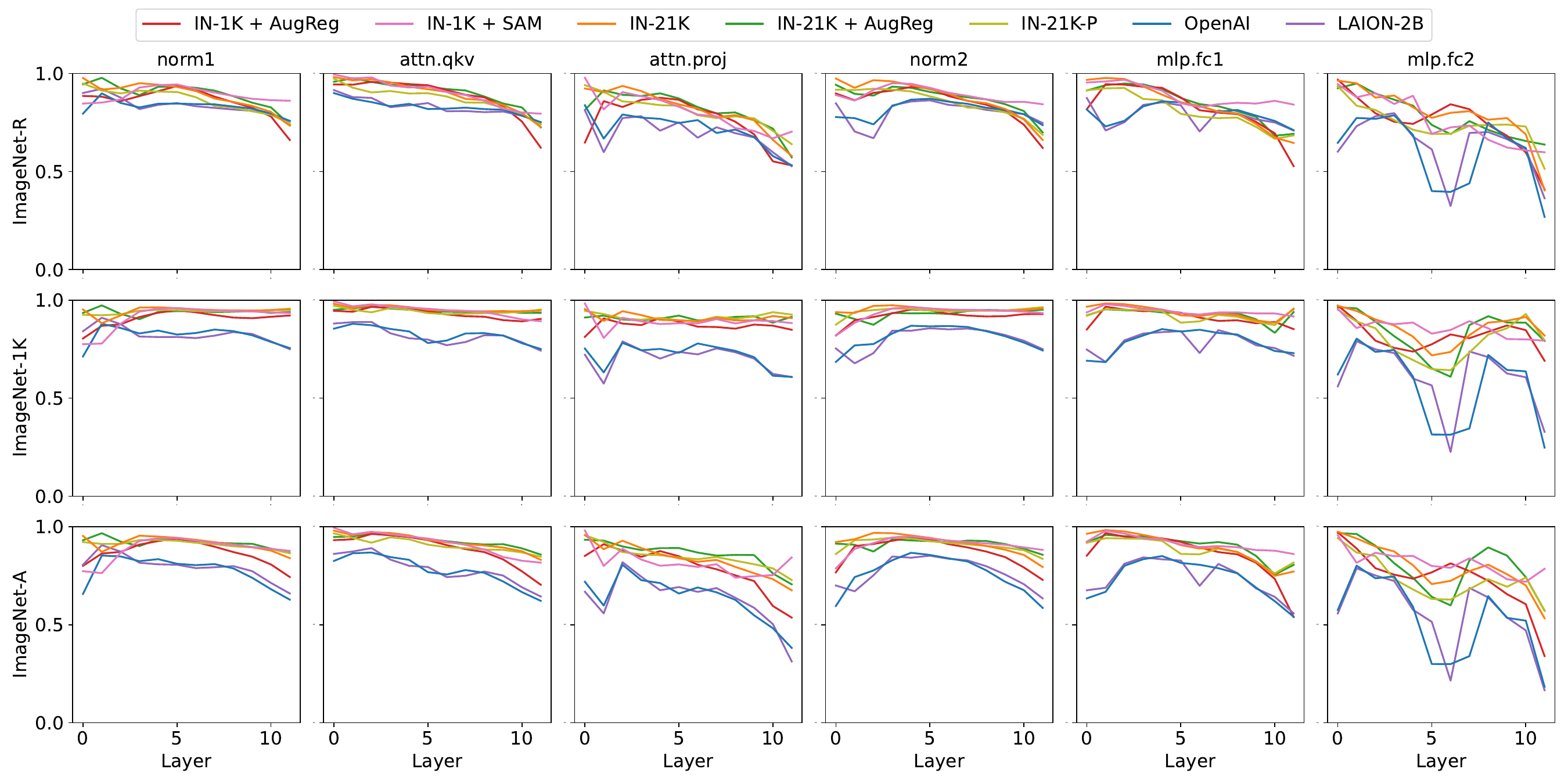}
\caption{
\textbf{Representational shifts from fine-tuning on ImageNet-R, analyzed by Centered Kernel Alignment (CKA), in ViT-B/16 models pretrained on various datasets.}
Rows indicate datasets where CKA is measured and column indicates different part of transformer demonstrated in Figure~\ref{fig:transformer}.
ImageNet-R (downstream), ImageNet-1K (pretraining), and ImageNet-A (OOD dataset) are used as evaluation datasets.
Models pretrained on OpenAI and LAION-2B show distinct CKA patterns across layers compared to others.
}
\label{fig:cka_imagenet_r}
\end{figure*}

\begin{table}[t!]
\caption{
Average accuracy on ImageNet-1K validation set and each OOD dataset using ViT-B/32 CLIP with zero-shot classifier before and after fine-tuning on the StanfordCars Dataset. Bold and denotes the best performance of pretrained and fine-tuned models, respectively.
}
\label{tab:stanford_cars}
\vspace{0.1cm}
\resizebox{\textwidth}{!}{
\begin{tabular}{c|c|ccccccccc}
\toprule
{\textbf{Pretraining Dataset}} & {\textbf{ImageNet-1K}} & {\textbf{AvgOOD}} & {\textbf{IN-V2}}         & {\textbf{IN-A}} & {\textbf{IN-R}} & {\textbf{IN-Sketch}} & {\textbf{ObjNet}} & {\textbf{IN-Cartoon}} & {\textbf{IN-Drawing}} & {\textbf{IN-C}} \\
\midrule
{LAION-100M (Pretrained)}      & {52.5}     & {38.6}& {44.5}       & {14.6}          & {64.5}          & {39.8}   & {30}  & {43.9}    & {\textbf{44.5}}       & {27.3}          \\
{LAION-100M (Fine-tuned)}      & {43.9}        & {28.6}   & {\textbf{36.2}} & {8.7}  & {53.9} & {29.7}      & {23.2}   & {34.1}       & {22.0}       & {20.8} \\
\midrule
{LAION-400M (Pretrained)}      & {59.6}     & {42.6}& {52.4}       & {19.6}          & {70.8}          & {46.4}   & {38.9}& {48.4}    & {29.4}    & {35.1}          \\
{LAION-400M (Fine-tuned)}      & {14.2}     & {9.2} & {12.5}       & {4.0}           & {19.6}          & {7.3}    & {9.3} & {9.8}     & {5.3}     & {5.8}           \\
\midrule
{OpenAI (Pretrained)}          & {60.2}     & {41.4}& {52.9}       & {28.3}          & {67.1}          & {40.4}   & {31.6}& {46.1}    & {29.1}    & {35.9}          \\
{OpenAI (Fine-tuned)}          & {5.7}      & {3.9} & {4.6}        & {2.1}           & {8.7}           & {2.3}    & {5.1} & {3.9}     & {2.1}     & {2.6}           \\
\midrule
{LAION-2B (Pretrained)}        & {\textbf{66.6}}        & {\textbf{48.7}}   & {\textbf{58.1}}          & {\textbf{26.3}} & {\textbf{76.4}} & {\textbf{53.7}}      & {\textbf{44.7}}   & {\textbf{55.8}}       & {33.2}    & {\textbf{41.6}} \\
{LAION-2B (Fine-tuned)}        & {6.8}      & {4.8} & {5.8}        & {2.1}           & {10.7}          & {3.7}    & {5.7} & {5.0}     & {2.5}     & {2.9}\\
\bottomrule
\end{tabular}
}
\end{table}

\subsection{Analysis of Representation Shift via Centered Kernel Alignment (CKA)}\label{subsec:cka}
To analyze intermediate representations before and after fine-tuning, we use Centered Kernel Alignment (CKA)~\citep{cortes2012algorithms,kornblith2019similarity}, the standard metrics to quantify similarity between neural network representations~\citep{kim2023stability,raghu2021vision}.
We fine-tune ViT-B/16 models pretrained on various datasets on ImageNet-R and measure the CKA between the pretrained and fine-tuned models on ImageNet-R (fine-tuning dataset), ImageNet-1K validation set, and ImageNet-A (OOD dataset).
Figure~\ref{fig:cka_imagenet_r} shows the CKA scores across transformer layers, broken down by component, as illustrated in Figure~\ref{fig:transformer}.
Deeper layers exhibit greater discrepancy between the pretrained and fine-tuned models.
Models pretrained on LAION-2B and OpenAI’s internal dataset show especially large discrepancies, particularly in earlier layers, compared to other models.
In addition, we observe a pronounced change in the \textsf{mlp.fc2} component of the sixth transformer block, in which pattern was observed in \citet{li2024surprising}.
This change becomes more significant in models pretrained on larger datasets, especially when evaluated on non-downstream datasets.
Considering the fact that this huge dissimilarity is not observed in the previous layer (\textsf{mlp.fc1}).
We expect that the sixth \textsf{mlp.fc2} layer may be linked to severe catastrophic forgetting, warranting further investigation.
This trend is maintained across various combinations of downstream datasets as shown in Figures~S1-S8 in the supplementary materials.
Freezing this layer during fine-tuning, however, does not prevent the collapse. Specifically, for the LAION-2B pretrained ViT-B/16, $mRI$ does not improve but instead worsens slightly, dropping from $-38.1$ under full fine-tuning to $-50.1$. This indicates that the shift at the sixth \textsf{mlp.fc2} is a correlate of the collapse rather than a localized cause that freezing can prevent, leaving it for future work.


\subsection{Stanford Car Dataset}\label{supp:stanford_cars}

We fine-tune various pretrained ViT-B/32 CLIPs on the StanfordCars dataset~\citep{krause20133d} with a zero-shot classifier (FT) and then evaluate their robustness on the full suite of OOD datasets used in the paper.
The dataset has 196 fine-grained car classes, unlike ImageNet variants.
As shown in Table~\ref{tab:stanford_cars}, the results align with our findings that a model pretrained on a larger dataset suffers more catastrophic forgetting.
The LAION-100M pretrained model does not suffer from severe performance degradation, whereas the other models do. This supports our hypothesis that pretraining on a large-scale dataset (LAION-400M, OpenAI, LAION-2B) is more likely to lead to severe catastrophic forgetting than pretraining on a smaller dataset (LAION-100M).

%% file: sections/5_discussion.tex

\section{Discussion}\label{sec:discussion}
\vspace{-0.2cm}

We found that models pretrained on larger, more diverse datasets, such as LAION-2B, experienced more severe robustness degradation after fine-tuning.
While these models exhibited high initial robustness, the performance drop was more prominent compared to models pretrained on smaller datasets like ImageNet-1K or LAION-100M, leading to even worse performance.
This degradation arises in the CLIP models and their fine-tuned variants and grows as their pretraining data scales, whereas the supervised models we evaluate do not show it (Section~\ref{subsec:paradox}), although its magnitude depends on the fine-tuning setting (Appendix~\ref{supp:optim_recipe}).
To facilitate these analyses, we introduced ImageNet-RIB (Robustness Inheritance Benchmark), a framework that evaluates model robustness across multiple downstream and OOD dataset pairs. In contrast to existing benchmarks that primarily consider a single downstream dataset~\citep{taori2020measuring}, ImageNet-RIB enables nuanced analyses of how varying fine-tuning contexts influence model generalization.
Moreover, we demonstrated that continual learning methods and robust fine-tuning approaches, particularly in combination, are effective in preserving or even improving robustness. 
Specifically, the combination of model soup with continual learning techniques consistently achieved superior performance.
This finding underscores the potential of integrating these strategies to mitigate catastrophic forgetting and enhance the robustness to OOD datasets.

Despite these contributions, our study has limitations.
Although we identify conditions under which extensive pretraining negatively impacts robustness and analyze feature representation, the underlying mechanisms remain unclear.
In particular, freezing the layer with the largest representational shift (the sixth \textsf{mlp.fc2}) during fine-tuning does not prevent the collapse (Section~\ref{subsec:cka}), suggesting the shift is a correlate rather than a localized cause.
Future research should investigate why extensive pretraining leads to worse robustness compared to smaller-scale pretraining, potentially informing more effective fine-tuning strategies.
Extending our evaluation to additional architectures and dataset contexts would further strengthen and generalize our conclusions.
In summary, our findings challenge common assumptions about pretraining dataset scale and robustness, emphasizing the importance of tailored fine-tuning strategies. We hope these insights motivate further investigation into optimizing robust fine-tuning practices, ultimately advancing the reliability and generalization capabilities of machine learning models.

%% file: sections/6_appendix.tex

\clearpage
\section*{Appendix}
We summarize each section in the Appendix as follows:

\paragraph{Appendix~\ref{supp:related} (Additional Related Works - Single Domain Generalization)} We survey single domain generalization 

\paragraph{Appendix~\ref{supp:distance} (Relationship between Dataset Distance and Accuracy Drop on Pretraining Dataset).} We quantify how distributional distance from ImageNet-1K relates to post-tuning performance.

\paragraph{Appendix~\ref{supp:small_set} (Fine-Tuning on Small Subset of Fine-Tuning Dataset).} We show that the robustness degradation happens when the dataset size is small.

\paragraph{Appendix~\ref{supp:split} (ImageNet-RIB with Train-Validation Split).} We split the fine-tuning dataset into a train set and a validation set and find the best-performing model on each setting.

\paragraph{Appendix~\ref{supp:effective} (Effective Robustness).} We demonstrate $mRI$ with effective robustness from the results in Appendix~\ref{supp:split}.

\paragraph{Appendix~\ref{supp:sec_exp_details} (Experimental Details).} We describe experimental details.

\paragraph{Appendix~\ref{supp:ablation} (Ablation Studies).} We conduct various ablation studies including learning rate, weight decay, multiple random seeds, and best ratio for WiSE-FT.

\paragraph{Appendix~\ref{supp:various} (Additional Experiments with Various Pretrained Models).} We include extensive experimental results such as robustness of pretrained models, backward transfer, performance on fine-tuning dataset, and full accuracies on experimental settings.

\section{Additional Related Works:  Single Domain Generalization}\label{supp:related}

Single-domain generalization refers to the task where only one source domain is available during training, and the model is evaluated on multiple unseen target domains~\citep{qiao2020learning}.
While the high-level concept is similar to the existing robust fine-tuning benchmark~\citep{taori2020measuring}, the objectives differ. Robust fine-tuning focuses on maintaining or improving a model's robustness to OOD datasets during fine-tuning, whereas single-domain generalization aims to achieve generalization to unseen OOD datasets, often through meta-learning-based data augmentation~\citep{chen2023meta,qiao2020learning} or adaptive batch normalization~\citep{fan2021adversarially}.
Recently, \citet{fan2021adversarially} apply single-domain generalization to the PACS dataset~\citep{li2017deeper}, using one domain as the training set and the remaining domains as test sets. This setup resembles our ImageNet-RIB benchmark in that each dataset is used for training while the others are used for testing. However, the goals of the two benchmarks differ: our robust fine-tuning benchmark aims to mitigate robustness degradation during fine-tuning, while single-domain generalization benchmarks focus on improving generalizability from a single source domain.

\begin{figure*}[t!]
    \centering
    \includegraphics[width=\linewidth]{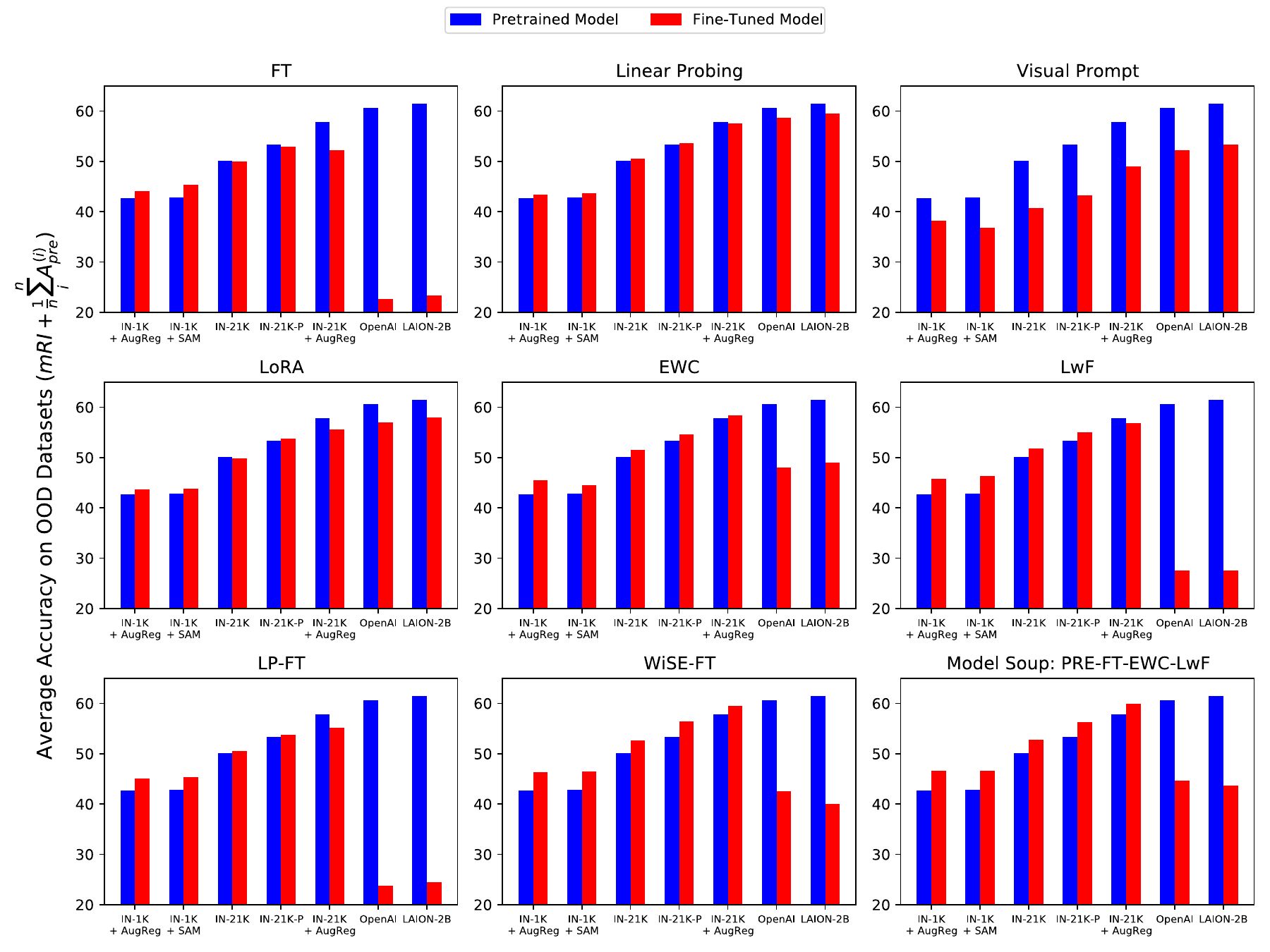}
\caption{
\textbf{The average accuracy on OOD datasets before (blue) and after (red) fine-tuning with each method on fine-tuning datasets.}
The red bar is calculated directly by evaluating pretrained models on OOD datasets while the blue bar is calculated by adding $mRI$ of each method to the pretrained models' accuracy.
Note that it is identical to the average accuracy on OOD datasets after fine-tuning on each dataset ($mRI + \frac{1}{n}\sum_i^n A^{(i)}_\text{pre} = \frac{1}{n}\sum_j \frac{1}{n-1}\sum_{i, i\neq j}^n A^{(i)}_\text{down}$).
Fine-tuning LAION-2B and OpenAI pretrained models on the fine-tuning datasets causes severe robustness loss leading to worse performance than ImageNet-1K with AugReg pretrained model.
Conversely, ImageNet-1K with the AugReg pretrained model improves robustness after fine-tuning.
Note that the difference between red and blue bars is $mRI$.
}
\label{fig:mri}
\end{figure*}


\begin{figure*}[t!]
    \centering
    \includegraphics[width=0.25\linewidth]{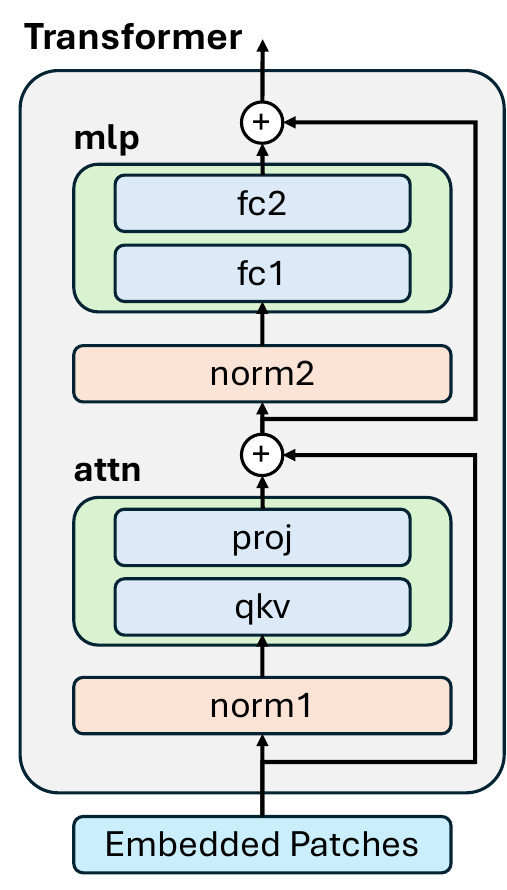}
\caption{
\textbf{Transformer Architecture of ViT-B/16.}
We describe only the weighted layers, \ie, fully connected layers and layer normalization layers.
}
\label{fig:transformer}
\end{figure*}


\section{Relationship between Dataset Distance and Accuracy Drop on Pretraining Dataset}\label{supp:distance}
\subsection{Optimal Transport Dataset Distance on Feature Space Aligns with Dataset Design Principles}
\begin{figure}[t!]
    \centering
    \begin{subfigure}[b]{0.4\textwidth}
        \centering
        \includegraphics[width=\textwidth]{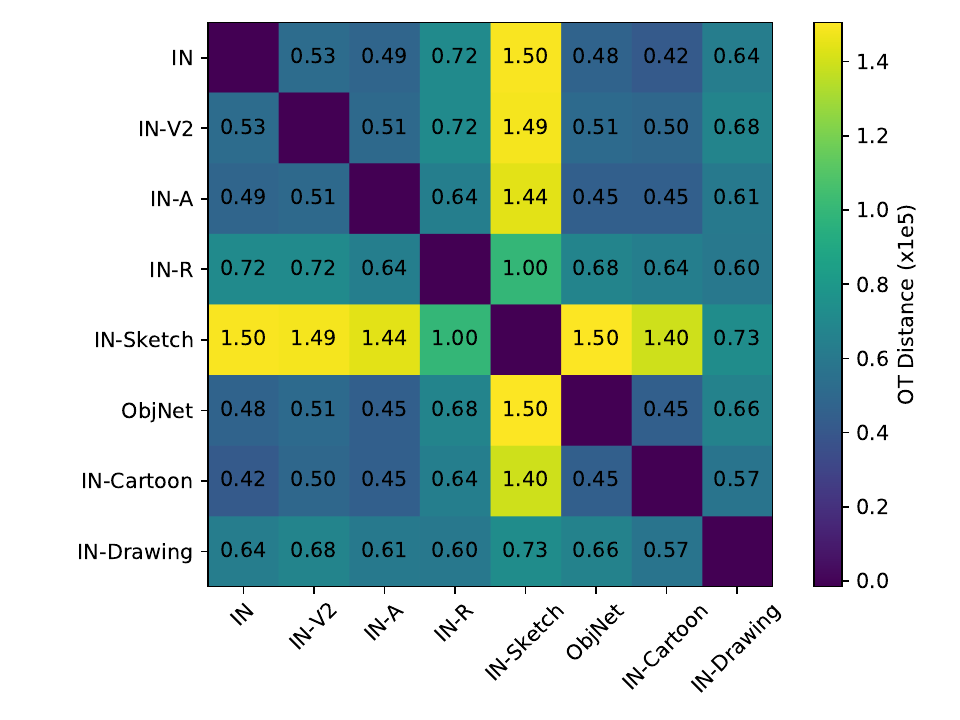}
        \caption{OTDD in the Image Space}
        \label{fig:distance:a}
    \end{subfigure}
    \begin{subfigure}[b]{0.4\textwidth}
        \centering
        \includegraphics[width=\textwidth]{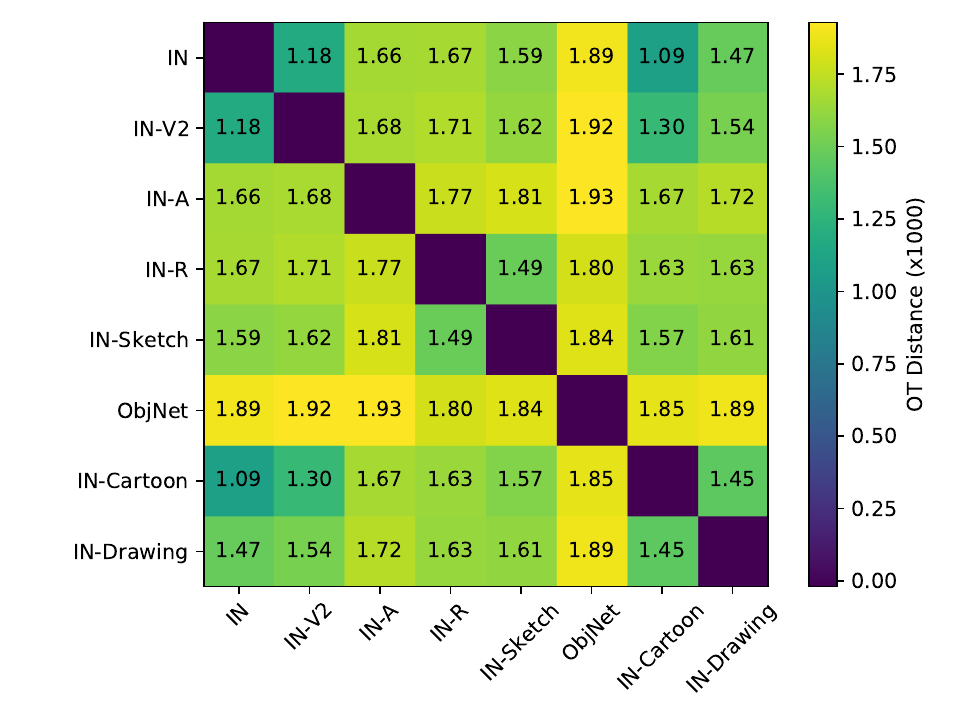}
        \caption{OTDD in the Feature Space}
        \label{fig:distance:b}
    \end{subfigure}

    \vskip\baselineskip

    \begin{subfigure}[b]{0.4\textwidth}
        \centering
        \includegraphics[width=\textwidth]{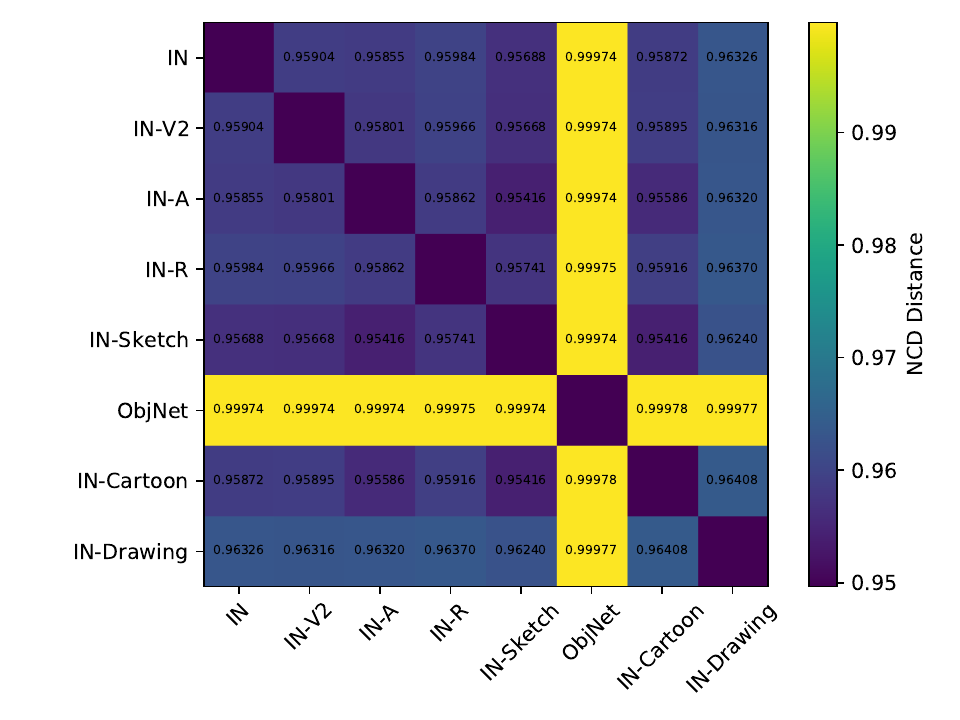}
        \caption{NCD in the Image Space}
        \label{fig:distance:c}
    \end{subfigure}
    \begin{subfigure}[b]{0.4\textwidth}
        \centering
        \includegraphics[width=\textwidth]{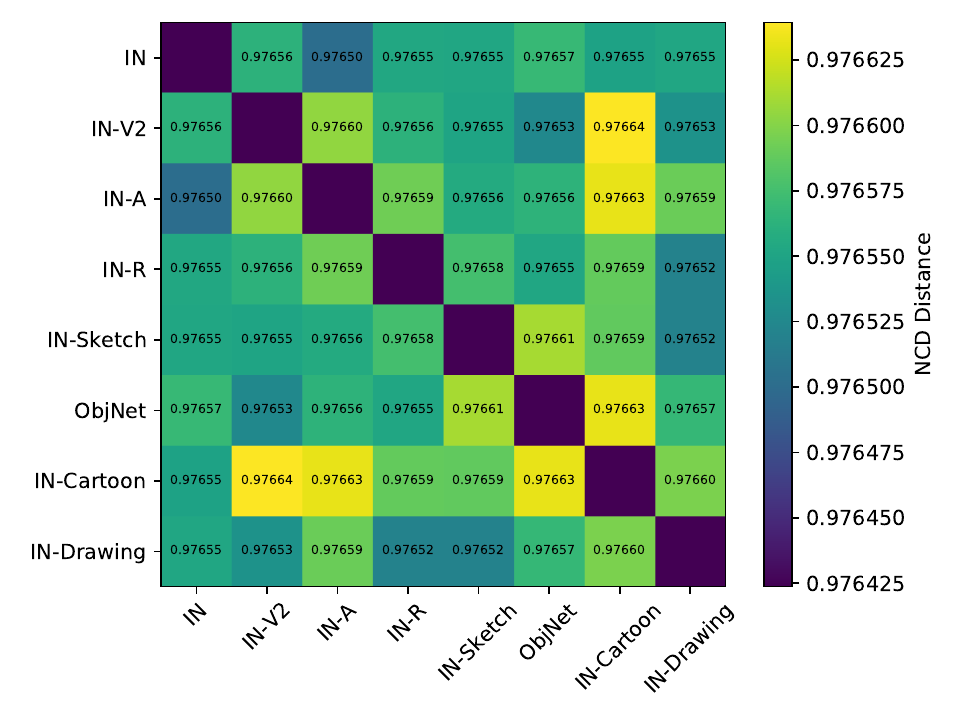}
        \caption{NCD in the Feature Space}
        \label{fig:distance:d}
    \end{subfigure}
    \caption{
    \textbf{Optimal Transport Dataset Distances (OTDD) in the feature space aligns with each dataset design.}
    Pairwise OTDD (up) and Normalized Compression Distance (NCD) (down) between datasets using images (left) and features extracted by ImageNet-1K with AugReg pretrained ViT-B/16 on each dataset (right), respectively.
    }
    \label{fig:distance}
\end{figure}

We measure the distance between datasets by using Optimal Transport Dataset Distance (OTDD)~\citep{alvarez2020geometric} and Normalized Compression Distance~(NCD)~\citep{cilibrasi2005clustering}.
The distance is measured in the image space and the feature space from ImageNet-1K with AugReg pretrained ViT-B/16, class tokens before the classifier layer.
Since ImageNet-C comprises multiple corruptions with different severities, we do not measure the distance to ImageNet-C. 
OTDD in the image space, ImageNet-Sketch is the farthest from other datasets as it is black and white sketch images~(Figure~\ref{fig:distance:a}).
ImageNet-Drawing is the closest to the dataset and the ImageNet-R is the second closest as they share the same styles and images, respectively.

OTDD in the feature space demonstrates a better alignment with the dataset design principles~(Figure~\ref{fig:distance:b}).
For example, ImageNet-V2 is designed to replicate the distribution of the ImageNet validation set.
It leads ImageNet-V2 the closest to ImageNet-1K among realistic datasets.
Moreover, the distances between ImageNet-1K and ImageNet-V2 to other datasets are consistent across both image and feature spaces.
This is not true with ImageNet-Cartoon since it is a synthetic dataset based on the ImageNet validation set.
 As shown in Table~\ref{tab:main_base_16_in1k}, ImageNet-Cartoon improves ImageNet-1K accuracy more than ImageNet-Drawing, suggesting that the distribution shift in cartoon-style images is less severe than that of drawing-style images.
Similarly, ObjectNet is intentionally collected with different viewpoints and backgrounds and it is the most distant from all other datasets in the feature space.

We also measure Normalized Compression Distance (NCD) using both images and the features from
ImageNet-1K with AugReg pretrained ViT-B/16. However, the distance between each dataset pair is
too insignificant to compare with each dataset as shown in Figures~\ref{fig:distance:c} and ~\ref{fig:distance:d}.

\begin{figure}[t!]
    \centering
    \includegraphics[width=\linewidth]{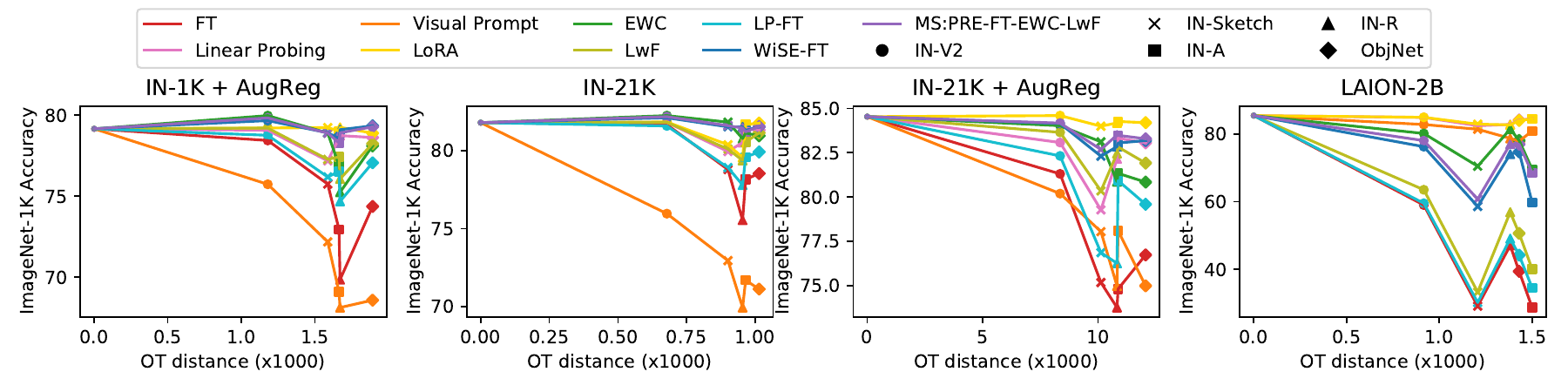}
    \vspace{-0.4cm}
\caption{
\textbf{Relationship between post fine-tuning ImageNet-1K accuracy and the distance between ImageNet-1K and the fine-tuning dataset.}
As the distance increases, accuracy generally decreases across fine-tuning methods. 
We exclude synthetic datasets made from ImageNet-1K validation set to avoid interference.
}
\label{fig:dist_acc}
\vspace{-0.4cm}
\end{figure}

\begin{table}[t!]
\caption{Pearson correlation coefficient between the accuracy on ImageNet-1K and the dataset distance between ImageNet-1K and each fine-tuning dataset. 
There is a negative correlation between accuracy and dataset distance.
Notably, FT and Prompter consistently exhibit a strong negative correlation across different pretrained models.}
\vspace{0.2cm}
\label{tab:corr}
    \centering
    \resizebox{0.9\columnwidth}{!}{
    \begin{tabular}{c|ccccccccc}
    \toprule
        Method & FT & LinearProbing & Visual Prompt & LoRA & EWC & LwF & LP-FT & WiSE-FT & Model Soup \\ \midrule
        IN-1K + AugReg & -0.64 & -0.22 & -0.91 & -0.63 & -0.57 & -0.49 & -0.59 & -0.46 & -0.54 \\ 
        IN-21K & -0.77 & -0.36 & -0.92 & -0.25 & -0.88 & -0.56 & -0.69 & -0.92 & -0.89 \\
        IN-21K + AugReg & -0.68 & 0.10 & -0.86 & -0.63 & -0.91 & -0.38 & -0.39 & -0.52 & -0.51 \\ 
        LAION-2B  & -0.67  & -0.19  & -0.74  & -0.31  & -0.32  & -0.44  & -0.56  & -0.31  & -0.13 \\
        \bottomrule
        
    \end{tabular}
    }
\end{table}

\subsection{Optimal Transport Dataset Distance Aligns With ImageNet-1K Accuracy Drop During Fine-Tuning}
We analyze how ImageNet-1K accuracy changes after fine-tuning on downstream datasets.
Using the Optimal Transport Dataset Distance (OTDD)~\citep{alvarez2020geometric} in the ViT-B/16 feature space, we find that accuracy generally decreases as OTDD from ImageNet-1K increases (Figure~\ref{fig:dist_acc}).
Pearson correlations (Table~\ref{tab:corr}) confirm a negative trend for all methods except linear probing, with FT and Visual Prompt showing strong correlations (< -0.5).
However, OTDD does not consistently correlate with out-of-distribution (OOD) accuracy post-fine-tuning.


\section{Fine-Tuning on Small Subset of Fine-Tuning Dataset}\label{supp:small_set}

\begin{figure*}[t!]
    \centering
    \includegraphics[width=0.75\linewidth]{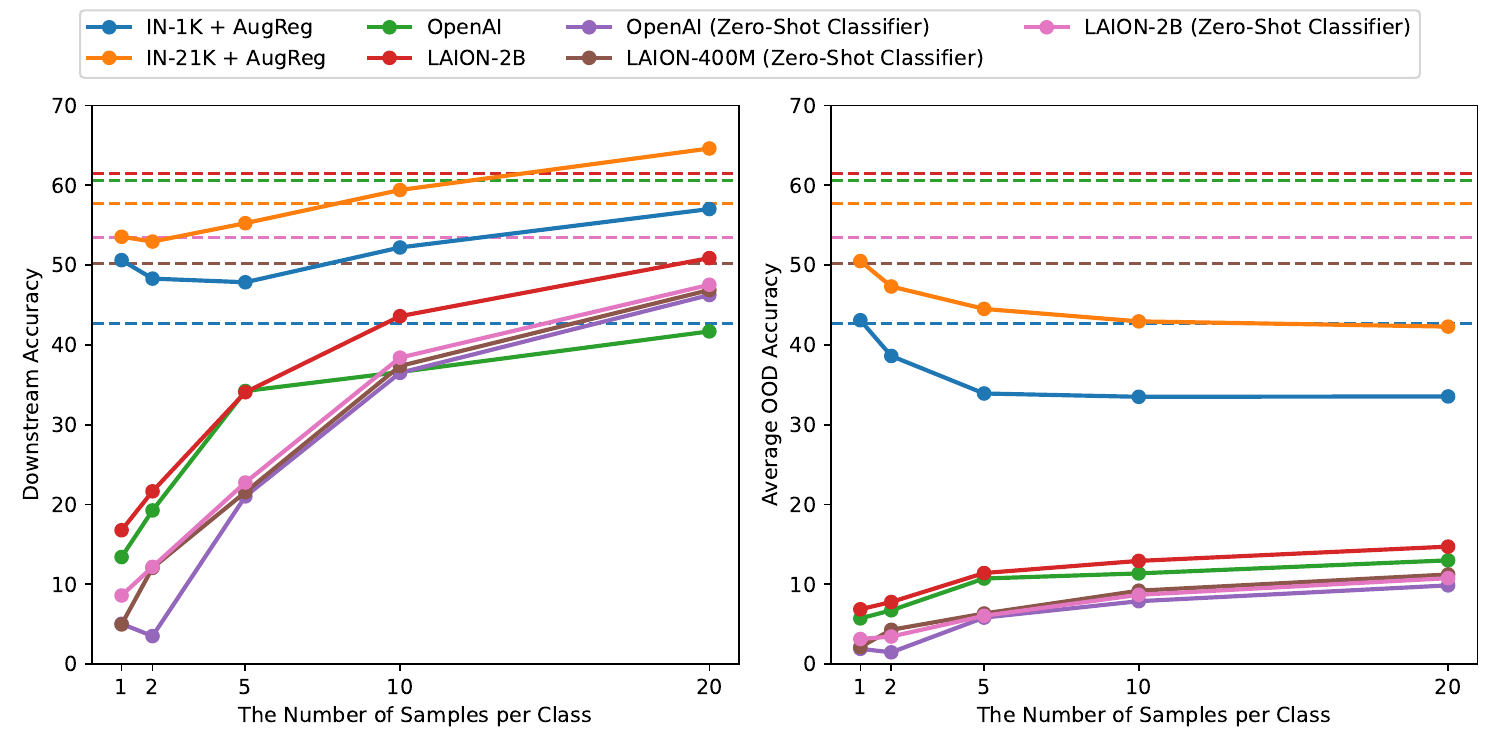}
\caption{
    \textbf{Fine-tuning on a small dataset leads to severe accuracy degradation in both in- and out-of-distribution.}  
Average accuracy on fine-tuning datasets and OOD datasets after fine-tuning pretrained ViT-B/16 models on a small number of images per class of fine-tuning datasets.
    The dashed line denotes the average accuracy of the pretrained models.
}
\label{fig:ft_small}
\end{figure*}

Figure~\ref{fig:ft_small} shows that CLIP models pretrained on large-scale datasets are particularly vulnerable when fine-tuned on small datasets.
To investigate whether this degradation is specific to CLIP or also affects classification models fine-tuned on ImageNet-1K, we compare an ImageNet-1K with AugReg and ImageNet-21K with AugReg pretrained ViT-B/16 models (classification models) with OpenAI and LAION-2B pretrained ViT-B/16 models, both in their original form and after classification fine-tuning.
We fine-tune these models on small subsets of each fine-tuning dataset within ImageNet-RIB and evaluate their average accuracy on both in-distribution (ID) and out-of-distribution (OOD) datasets.
While ImageNet-21K with AugReg pretrained model also experiences a performance drop on the fine-tuning dataset when the number of samples per class falls below 10, the degradation is significantly less severe than that observed for LAION-2B models.
On the other hand, ImageNet-1K with AugReg pretrained model's performance increases.
As the fine-tuning dataset size increases, the OOD accuracy of classification models gradually declines, indicating a progressive adaptation to the specific dataset. 
In contrast, the OOD accuracy of CLIP models and ones fine-tuned on ImageNet-1K initially collapse but then increase with more fine-tuning, suggesting a different adaptation mechanism.

\begin{table}[t!]
\centering
\caption{$mRI$ values obtained using the best validation accuracy for each model on the fine-tuning datasets. Parentheses indicate the accuracy difference compared to models fine-tuned for 10 epochs without splitting the training and validation sets.
The average number of epochs needed to achieve the highest validation accuracy on each fine-tuning dataset. 
$\dagger$: WiSE-FT and Model Soup are post-hoc weight interpolation methods and do not involve training.
}\vspace{0.1cm}
\label{tab:mri_val}
\resizebox{0.9\textwidth}{!}{
\begin{tabular}{c|cc|cc|cc}
\toprule
\multirow{2}{*}{Method} &  \multicolumn{2}{c|}{IN-1K + AugReg}   &\multicolumn{2}{c|}{IN-21K + AugReg}   & \multicolumn{2}{c}{LAION-2B} \\
& $mRI$ & Best Epoch & $mRI$ & Best Epoch & $mRI$ & Best Epoch \\
\midrule
FT    & 2.8 ({\color{red}+1.5}) & 4.2    & -5.6 ({\color{blue}-0.1})  &  7.1   & -40.4 ({\color{blue}-2.3}) &  12.1  \\
Linear Probing   & 1.6 ({\color{red}+0.9})   & 17.8 & -0.7 ({\color{blue}-0.4}) & 11.5        & \textbf{-2.3 ({\color{blue}-0.3})} & 14.1 \\
Visual Prompt    & -4.1 ({\color{red}+0.4}) & 22.4 &  -9.1 ({\color{blue}-0.3}) & 22.4       & -8.9 ({\color{blue}-0.7}) & 20.6   \\
LoRA & 2.6 ({\color{red}+1.7})  &20.1 &   1.2 ({\color{red}+3.3})  & 16.5         & -2.9 ({\color{red}+0.7}) & 15.6  \\
EWC   & 3.6 ({\color{red}+0.8})   & 18.1 &  0.5 ({\color{blue}-0.1})  & 19.8        & -14.1 ({\color{blue}-1.6}) & 24.5  \\
LwF & 4.0 ({\color{red}+0.9})  & 3.8 &   -1.4 ({\color{blue}-0.4}) & 4.2        & -34.6 ({\color{blue}-0.7})  & 11.5  \\
LP-FT& 3.7 ({\color{red}+1.4})  & 7.2 &  -2.4 ({\color{red}+0.2}) & 13.2        & -36.7 ({\color{red}+0.4}) & 13.8   \\
WiSE-FT & 4.5 ({\color{red}+0.9})    & -$^\dagger$  & 1.6 ({\color{blue}-0.1}) &-   & -23.0 ({\color{blue}-1.4}) & - \\
MS:PRE-FT-EWC-LwF & \textbf{4.8 ({\color{red}+0.9})} & - & \textbf{2.0 ({\color{blue}-0.2})} & - &  -19.2 ({\color{blue}-1.3})& - \\
\bottomrule
\end{tabular}
}
\end{table}

\section{ImageNet-RIB with Train-Validation Split}\label{supp:split}
In the original ImageNet-RIB benchmark, the entire fine-tuning dataset is used for fine-tuning.
 To evaluate the robustness of the models under a different setup, we introduce a train-validation split (4:1 ratio), where models are fine-tuned on the training set, validated on the validation set, and then evaluated on the benchmark using the best-performing epoch from the validation set.
We divide the fine-tuning dataset into training and validation sets, extending the training duration to 25 epochs to identify the optimal model based on validation accuracy.
Using the best-performing model, we applied robust fine-tuning methods, including LP-FT, WiSE-FT, and Model Soup, to evaluate out-of-distribution (OOD) performance.

For this variant of the benchmark, we showcase results using ViT-B/16 models pretrained on ImageNet-1K with AugReg, ImageNet-21K with AugReg, and LAION-2B.
Table~\ref{tab:mri_val} reports the mean Robustness Improvement ($mRI$) and the average number of epochs required for each model to achieve its highest validation accuracy on the fine-tuning datasets.
Notably, the performance under the train-validation split does not significantly differ from the results in Table~\ref{tab:mri}, where the entire fine-tuning dataset is used for fine-tuning with models trained for 10 epochs.

\section{Effective Robustness}\label{supp:effective}

As we mentioned in Section~\ref{sec:benchmark}, our robust improvement metrics is based on relative robustness.
In this section, we employ effective robustness~\citep{taori2020measuring} instead of direct accuracy difference (relative robustness) to compute effective robustness improvement, $eRI$~(mean effective robustness while fine-tuning on dataset $D_i$):
\begin{equation}
	eRI_i = \frac{1}{n-1} \sum_{j=1, j \neq i}^n A^{(j)}_i - \beta_j(A_i^{(i)}),
\end{equation}
where $\beta_j(x)$ denotes a baseline accuracy on dataset $D_j$ when the accuracy on the dataset $D_i$ is $x$.
We calculate mean $eRI$ in ImageNet-RIB with Train-Validation Split (Appendix~\ref{supp:split}).
Note that in the original ImageNet-RIB setting, we use the entire dataset for training, not splitting the validation set. 
Table~\ref{tab:effective} demonstrates that all mean $eRI$ become negative.
 This is because, unlike previous benchmark~\citep{shi2023effective,taori2020measuring} where the downstream dataset (ImageNet-1K) contains all labels in OOD datasets, our downstream dataset does not contain all labels (\eg, ImageNet-R has 200 classes while ImageNet-Sketch has 1000 classes).
That is why the robustness can be negative. Even under this condition, the LAION-2B retrained model with FT performs much worse than others.

\begin{table}[t!]
\centering
\caption{
mean effective robustness improvement across OOD datasets obtained using the best validation accuracy for each model on the fine-tuning datasets. Parentheses indicate the accuracy difference compared to models fine-tuned for 10 epochs without splitting the training and validation sets.
The average number of epochs needed to achieve the highest validation accuracy on each fine-tuning dataset. 
}
\label{tab:effective}
\vspace{0.1cm}
\resizebox{0.7\textwidth}{!}{
\begin{tabular}{c|c|c|c}
\toprule
Method & IN-1K + AugReg & IN-21K + AugReg & LAION-2B\\
\midrule
FT & -17.6 & -18.5 & -31.7\\
linear Probing & -11.4 & -9.2 & -9.7\\
Visual Prompt & -13.6 & -12.8 & -10.6\\
LoRA & \textbf{-3.7} & \textbf{-5.4} & \textbf{-0.3}\\
EWC	  &  -11.5 & -11.2 & -19.5\\
LwF	 &  -15.3 & -13.4 & -29.3\\
LP-FT & -17.3 & -16.4 & -30.9\\
Wise-FT & -11.6 & -11.6 & -21.9\\
MS:PRE-FT-EWC-LwF & -12.7 & -12.0 & -21.4\\
\bottomrule
\end{tabular}
}
\end{table}


\section{Experimental Details}\label{supp:sec_exp_details}
In this section, we describe the details of the experimental setup.
We use a single NVIDIA RTX 4090 GPU for the experiment.

\subsection{Out-of-Distribution Datasets in ImageNet-RIB}~\label{supp:dataset}

We leverage all existing ImageNet variants designed to measure the robustness of the trained network during distribution shifts.
ImageNet-O~\citep{hendrycks2021natural} is not used since it is an out-of-distribution detection dataset.

\paragraph{ImageNet-V2~\citep{recht2019imagenet}}
ImageNet-V2 is designed to have a distribution as similar as possible to the original ImageNet-1K.
It has 50,000 images with 1,000 classes same as the original validation set.
The dataset is used under the MIT license.

\paragraph{ImageNet-A~\citep{hendrycks2021natural}}
ImageNet-A is an adversarially filtered test image that ImageNet-1K pretrained ResNet-50~\citep{he2016deep} is difficult to predict correctly.
It contains 7,500 images with 200 difficult subclasses from ImageNet-1K.
The dataset is used under the MIT license.

\paragraph{ImageNet-R~\citep{hendrycks2021many}}
ImageNet-R (Renditions) contains 30,000 images from 200 ImageNet classes with various rendition styles such as painting, sculpture, embroidery, origami, cartoon, toy, and so on.
The drawing rendition overlaps with ImageNet-Sketch~\citep{wang2019learning}.
The dataset is used under the MIT license.

\paragraph{ImageNet-Sketch~\citep{wang2019learning}}
ImageNet-Sketch comprises black and white sketch drawings of the ImageNet-1K classes and each class has 50 images.
The dataset is used under the MIT license.

\paragraph{ImageNet-Cartoon and ImageNet-Drawing~\citep{salvador2022imagenetcartoon}}
ImageNet-Cartoon and ImageNet-Drawing are to be converted from ImageNet validation set images to cartoon, and drawing styles based on generative adversarial network~\citep{wang2020learning} and image processing~\citep{lu2012combining}.
These simplified representations test a model's ability to identify objects from minimalistic and abstract visual information.
The dataset is used under the Creative Commons Attribution 4.0 International license.

\paragraph{ObjectNet~\citep{barbu2019objectnet}}
ObjectNet is designed for evaluating object recognition models under more realistic conditions such as various poses, backgrounds, and viewpoints.
There are 50,000 images with 313 object classes and 113 classes are overlapped with ImageNet.
We only use ImageNet class objects. 
The dataset is used under the MIT license.

\paragraph{ImageNet-C~\citep{hendrycks2019benchmarking}}
ImageNet-C is designed for measuring the robustness of models to common perturbations such as noise, blur, weather, and digital distortions.
In the dataset, ImageNet validation set images are perturbed with various severity from 1 to 5.
Unlike the original metrics, corruption error compared with AlexNet, we use average accuracy for consistency with other datasets.
The dataset is used under the Apache-2.0 license.

\begin{table}[t!]
\centering
\caption{
Python libraries and the names of network weights for each pretrained model.
 }
\label{tab:library}
\resizebox{0.8\columnwidth}{!}{
\begin{tabular}{c|c|c|c}
\toprule
Architecture & Pretraining Dataset & Library & Weight Name\\
\midrule
\multirow{7}{*}{ViT-B/16} & IN-1K + AugReg & \textsf{timm} & \textsf{vit\_base\_patch16\_224.augreg\_in1k}\\
 & IN-1K + SAM & \textsf{timm} & \textsf{vit\_base\_patch16\_224.sam\_in1k}\\
 & IN-21K&  \textsf{timm} & \textsf{vit\_base\_patch16\_224.orig\_in21k\_ft\_in1k}\\
 & IN-21K + AugReg& \textsf{timm}		& \textsf{vit\_base\_patch16\_224.augreg\_in21k\_ft\_in1k}\\
 & IN-21K-P & \textsf{timm}		& \textsf{vit\_base\_patch16\_224\_miil.in21k\_ft\_in1k}\\
 & LAION-2B& \textsf{timm} 	& \textsf{vit\_base\_patch16\_clip\_224.laion2b\_ft\_in1k}\\
  & OpenAI & \textsf{timm} 	& \textsf{vit\_base\_patch16\_clip\_224.openai\_ft\_in1k}\\
\midrule
\multirow{4}{*}{ViT-B/32} & IN-1K + AugReg & 	\textsf{timm}	& \textsf{vit\_base\_patch32\_224.augreg\_in1k} \\
 & IN-21K  + AugReg	&	\textsf{timm}		& \textsf{vit\_base\_patch32\_224.augreg\_in21k\_ft\_in1k} \\
 & LAION-2B& \textsf{timm} &	\textsf{vit\_base\_patch32\_clip\_224.laion2b\_ft\_in1k}\\
  & OpenAI & \textsf{timm} &	\textsf{vit\_base\_patch32\_clip\_224.openai\_ft\_in1k}\\
\midrule
\multirow{2}{*}{ViT-S/16} & IN-1K + AugReg& \textsf{timm} & \textsf{vit\_small\_patch16\_224.augreg\_in1k}\\
& IN-21K  + AugReg& \textsf{timm} & \textsf{vit\_small\_patch16\_224.augreg\_in21k\_ft\_in1k}\\
\midrule
ViT-S/32 & IN-21K  + AugReg& \textsf{timm} & \textsf{vit\_small\_patch32\_224.augreg\_in21k\_ft\_in1k}\\
\midrule
\multirow{1}{*}{ViT-L/16}& IN-21K  + AugReg& \textsf{timm} & \textsf{vit\_large\_patch16\_224.augreg\_in21k\_ft\_in1k}\\
\midrule
\multirow{1}{*}{ResNet-18} & IN-1K  & \textsf{torchvision} &\textsf{ResNet18\_Weights.DEFAULT}\\
\midrule
\multirow{1}{*}{ResNet-50} & IN-1K & \textsf{torchvision} &\textsf{ResNet50\_Weights.DEFAULT}\\
\bottomrule
\end{tabular}
}
\end{table}

\subsection{Pretrained Model}\label{supp:weight}

Table~\ref{tab:library} lists the libraries and corresponding network weight names for each model. 
We use the entire models in \textsf{timm} and \textsf{torchvision} library, which are finally fine-tuned on ImageNet-1K, with patch sizes of 16 and 32, and input image shape of 224 among ViT small, base, and large.
For ResNets, we use the default ImageNet-1K pretrained weights from the \textsf{torchvision} library.

\subsection{Training and Hyperparameters}\label{supp:exp_details}
Each pretrained model is fine-tuned on the downstream dataset for 10 epochs where the average accuracy on fine-tuning datasets for each pretrained ViT-B/16 model achieves more than 90\% with vanilla fine-tuning.
We applied LoRA on query and value projection layers with rank 8 following the original implementation~\citep{hu2021lora}.
We use 2 as a temperature for calculating KL divergence for LwF following \citet{li2017learning}.
For WiSE-FT, we use the interpolation ratio between pretrained and fine-tuned models as $0.5$ following the recommendation by \citet{wortsman2022robust} instead of finding the best hyperparameters evaluated on the benchmark for the fair comparison.
Appendix~\ref{supp:wiseft} compares with results of the best-performing ratio.

\section{Ablation Studies}\label{supp:ablation}

\subsection{Catastrophic Forgetting Persists Despite Small Learning Rates}\label{supp:lr}
\begin{figure*}[t!]
    \centering
    \includegraphics[width=0.75\linewidth]{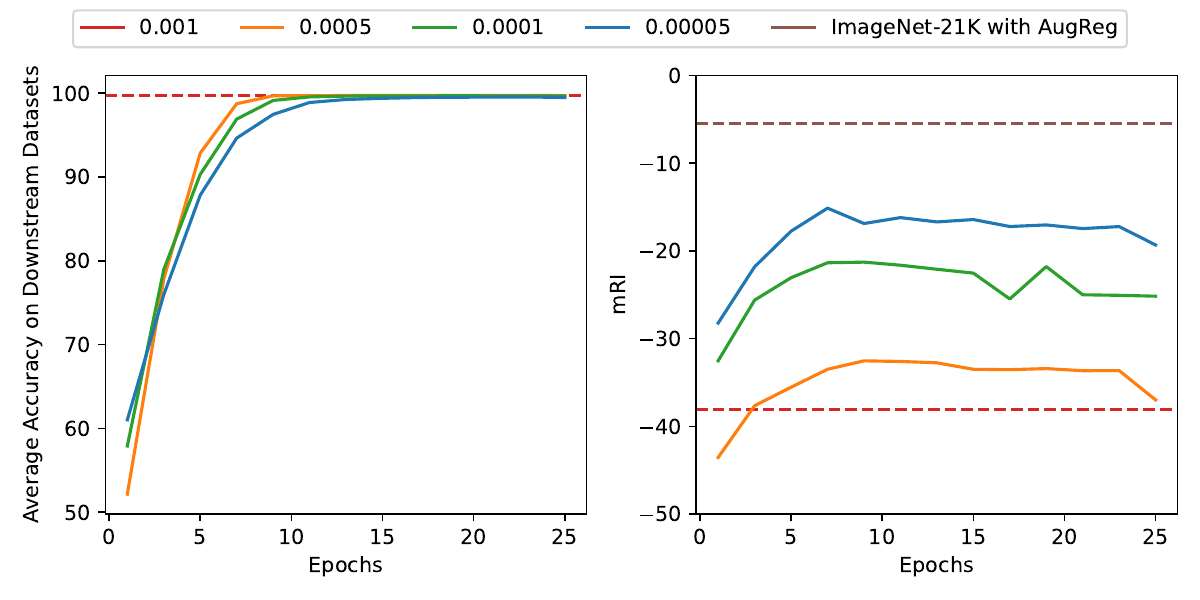}
\caption{
    \textbf{$mRI$ is much lower than ImageNet-21K with AugReg model, even with a small learning rate.}  
Average accuracy on the fine-tuning datasets and $mRI$ while fine-tuning pretrained ViT-B/16 models.
}
\label{fig:ablation_lr}
\end{figure*}

To test whether catastrophic forgetting stems solely from large learning rates, we fine-tune the LAION-2B pretrained ViT-B/16 on fine-tuning datasets using reduced learning rates (0.0005, 0.0001, 0.00005).
We extend training to 25 epochs to account for slower learning rates and evaluate performance throughout training (Figure~\ref{fig:ablation_lr}).
Although a smaller learning rate attenuates forgetting, once the fine-tuning dataset is fit, $mRI$ remains substantially below that of the ImageNet-21K + AugReg baseline ($-5.5$; Table~\ref{tab:mri}).
This implies that a smaller learning rate alone is not a solution for the severe robustness degradation.

\subsection{Weight Decays}\label{supp:weight_decay}
Beyond the learning rate analysis mentioned in Appendix~\ref{supp:lr}, we also fine-tune LAION-2B pre-trained VIT-B/16 with various weight decays in Table~\ref{tab:weight_decay}.
Across all settings, $mRI$ is substantially lower than ViT-B/16 pretrained on smaller datasets in Table~\ref{tab:mri}.

\begin{table}[t!]
\caption{$mRI$ of ViT-B/16 pre-trained on LAION-2B with various weight decay used in fine-tuning.
We use vanilla fine-tuning (FT).}
\label{tab:weight_decay}
\centering
\resizebox{0.7\textwidth}{!}{
\begin{tabular}{c|ccccccc}
\toprule
Weight Decay &0 & 0.0001 & 0.0005 & 0.001 & 0.005 & 0.01 & 0.1   \\
\midrule
$mRI$  & \textbf{-38.1} & -39.1 & -44.2 & -39.4 & -40.9 & -49.4 & -59.3 \\
\bottomrule
\end{tabular}
}
\end{table}

\subsection{Optimizer}\label{supp:optim_recipe}
The main paper fine-tunes every model with SGD with learning rate 0.001, the standard setting for ImageNet classifiers~\citep{dosovitskiy2021image,steiner2022how}, which every backbone in our study is before $\Ddown$ fine-tuning (Section~\ref{subsec:exp_details}).
Robust fine-tuning methods for CLIP models such as WiSE-FT~\citep{wortsman2022robust} and FLYP~\citep{goyal2023finetune} instead start from the pretrained CLIP model with its zero-shot classifier and use AdamW with learning rates on the scale of 0.00001.
Table~\ref{tab:recipe_ablation} therefore repeats the ViT-B/16 experiments for four pretrainings fine-tuned on ImageNet-1K as in Table~\ref{tab:mri} and three zero-shot CLIP models as in Table~\ref{tab:zeroshot}, under AdamW with learning rates 0.001 and 0.00001 and the protocol of Section~\ref{sec:exp}.
The values under SGD with learning rate 0.001 are the FT rows of those two tables.
Under AdamW with learning rate 0.00001 the collapse is much milder, but the ordering persists; the ImageNet-supervised models stay within three points of zero, whereas LAION-2B and OpenAI still incur the largest degradations.
AdamW at learning rate 0.001 leads catastrophic forgetting, and the CLIP models do not even fit the fine-tuning dataset, whereas SGD with a smaller learning rate only delays the forgetting (see Appendix~\ref{supp:lr}).
The severity of the collapse thus depends on the optimizer and learning rate together, while the ordering across pretraining datasets does not.

\begin{table}[t!]
\centering
\caption{$mRI$ and mean accuracy on the fine-tuning dataset (Down.Acc) of ViT-B/16 under AdamW with learning rates 0.001 and 0.00001, for four pretrainings fine-tuned on ImageNet-1K and three zero-shot CLIP models.
Only AdamW with learning rate 0.00001 both fits the target and retains robustness.}
\label{tab:recipe_ablation}
\vspace{-0.2cm}
\resizebox{0.7\textwidth}{!}{
\begin{tabular}{c|c|cc|cc}
\toprule
\multirow{2}{*}{Model} & \multirow{2}{*}{Pretraining} & \multicolumn{2}{c|}{AdamW-0.001} & \multicolumn{2}{c}{AdamW-0.00001} \\
 & & $mRI$ & Down.Acc & $mRI$ & Down.Acc \\
\midrule
\multirow{4}{*}{\shortstack[c]{Fine-tuned on\\ImageNet-1K}} & IN-1K + AugReg & -30.5 & 99.7 & 1.7 & 97.1 \\
 & IN-21K + AugReg & -49.2 & 99.7 & -2.2 & 99.7 \\
 & OpenAI & -58.6 & 49.9 & -5.7 & 99.7 \\
 & LAION-2B & -56.1 & 78.7 & -4.8 & 99.7 \\
\midrule
\multirow{3}{*}{Zero-shot CLIP} & 
LAION-400M & -49.2 & 27.9 & -2.2 & 99.6 \\
& OpenAI & -47.7 & 29.3 & -0.9 & 99.6 \\
& LAION-2B & -52.1 & 51.1 & -2.6 & 99.6 \\
\bottomrule
\end{tabular}
}
\vspace{-0.2cm}
\end{table}

\subsection{Robustness to Multiple Random Seeds}
We test whether the results vary depending on the random seeds.
Table~\ref{tab:seed} illustrates $mRI$ of three different pretrained ViT-B/16.
As the data clearly shows, the standard errors across all runs are very small.
This confirms that our findings are highly stable and not an artifact of a particular random seed. Most importantly, the central conclusions of our paper are fully supported by this statistical analysis.
For the vanilla fine-tuning method, there remains a massive gap in post-tuning robustness (mRI) between the LAION-2B model and the models pretrained on ImageNet.

\begin{table}[t!]
\caption{$mRI$ of ViT-B/16 pre-trained on three different datasets after fine-tuning with different fine-tuning methods. We run with three random seeds and calculate the mean and standard error.}
\label{tab:seed}
\centering
\resizebox{0.9\textwidth}{!}{
\begin{tabular}{c|cccc}
\toprule
Method & ImageNet 1K + AugReg & ImageNet 21K + AugReg & LAION-2B    \\
\midrule
FT     & 1.4 $\pm$ 0.0& -5.4 $\pm$ 0.1& -38.2 $\pm$ 0.0 \\
Linear Probing     & 0.8 $\pm$ 0.0& -0.3 $\pm$ 0.0& -2.0 $\pm$ 0.1  \\
EWC    & 2.9 $\pm$ 0.1& 0.8 $\pm$ 0.1 & -12.9 $\pm$ 0.4 \\
LwF    & 3.2 $\pm$ 0.1& -0.8 $\pm$ 0.1& -33.5 $\pm$ 0.2 \\
LP-FT  & 2.0 $\pm$ 0.2& -2.7 $\pm$ 0.2& -37.2 $\pm$ 0.1 \\
Wise-FT& 3.7 $\pm$ 0.1& 1.9 $\pm$ 0.1 & -21.5 $\pm$ 0.3 \\
MS: PRE-FT-EWC-LwF & 4.0 $\pm$ 0.1& 2.3 $\pm$ 0.1 & -18.0 $\pm$ 0.2\\
\bottomrule
\end{tabular}
}
\end{table}

\subsection{The Best  Ratio for WiSE-FT}\label{supp:wiseft}
We conduct a grid search from 0.1 to 0.9 with an increment of 0.1 to find the best-performing ratio ($\alpha$) between the pretrained ViT-B/16's weight and the fine-tuned model’s weight for WiSE-FT:
\begin{equation}
\Wbf_\text{WiSE-FT} = \alpha \cdot \Wbf_\text{pre} + (1-\alpha)\cdot \Wbf_\text{FT},
\end{equation}
where $\Wbf_\text{WiSE-FT}$, $\Wbf_\text{pre}$, and $\Wbf_\text{FT}$ represent the network weights of WiSE-FT, the pretrained model, and the vanilla fine-tuned model, respectively.

It is important to note that this hyperparameter search, based on test results ($mRI$), constitutes an unfair comparison with other methods.
Table~\ref{tab:wise_ratio} compares the $mRI$ achieved by WiSE-FT using the default ratio of 0.5 from WiSE-FT~\citep{wortsman2022robust} and the best ratio.
WiSE-FT using OpenAI or LAION-2B pretrained models performs significantly better with the best ratio, as it relies minimally on the fine-tuned model’s weights.
 Similarly, hyperparameter search for Model Soup, which combines network weights from pretrained and fine-tuned models (e.g., FT, EWC, LwF), could further improve performance. 
 Notably, WiSE-FT is a special case of Model Soup when the ratios for EWC and LwF are set to 0.
However, exploring the optimal ratio for Model Soup weights is beyond the scope of this study.

\begin{table}[t!]
\centering
\caption{$mRI$ of ViT-B/16 pretrained on various datasets using WiSE-FT with default ratio (0.5) and the best ratio
}
\label{tab:wise_ratio}
\resizebox{0.75\textwidth}{!}{
\begin{tabular}{c|ccc}
\toprule
Pretraining Dataset & WiSE-FT ($\alpha=0.5$) & WiSE-FT (best $\alpha$) & best $\alpha$\\
\midrule
IN-1K + AugReg& 3.6	&4.7	&0.4\\
IN-1K + SAM&	3.6	&4.1	&0.3\\
IN-21K&	2.5	&2.5	&0.5\\
IN-21K-P&	3.0&	3.0	&0.5\\
IN-21K + AugReg&	1.7&	2.3	&0.7\\
OpenAI&	-18.1 &	-1.6&	0.9\\
LAION-2B	&-21.6&	-2.4	&0.9\\
\bottomrule
\end{tabular}
}
\end{table}

\section{Additional Experiments with Various pretrained Models}\label{supp:various}

\subsection{Robustness of pretrained Models}\label{supp:models}
We evaluate pretrained models mentioned in Appendix~\ref{supp:weight} on OOD datasets as shown in Table~\ref{tab:pre}.
Larger networks with smaller patch sizes achieve higher accuracy on both ImageNet-1K and OOD datasets.
Similarly, models pretrained on larger, more diverse datasets demonstrate better performance.

\input{method_first_tables/pre}

\input{method_first_tables/backward_transfer}

\subsection{Backward Transfer}
We measure the backward transfer, accuracy change on the pretraining dataset, ImageNet-1K validation set after fine-tuning.
Table~\ref{tab:backward_transfer} presents the average backward transfer across different fine-tuning methods on downstream datasets.
While LwF achieves the best backward transfer in most models, linear probing outperforms it on OpenAI and LAION-2B pretrained models.

\subsection{Performance on Fine-Tuning Dataset}\label{supp:downstream}

Tables~\ref{tab:down_base} ~\ref{tab:down_other}, and ~\ref{tab:down_res} demonstrate the accuracy on fine-tuning datasets~(\ie, training accuracy) with ViT base, ViT large and ViT small, and ResNet, respectively.
FT, LwF, and LP-FT can overfit to the fine-tuning dataset but WiSE-FT and Model Soup (PRE-FT-LwF-EWC) have worse performance which might be due to using pretrained model weights.
Visual Prompt and LoRA rarely learn from a fine-tuning dataset.

\input{method_first_tables/down_base}
\input{method_first_tables/down_other}
\input{method_first_tables/down_res}

\clearpage

\subsection{Robustness Improvement Results of Different Models}
Across the ImageNet pretrained models, WiSE-FT and Model Soup consistently have better robustness improvement compared to other methods fine-tuning on realistic OOD datasets (Tables~\ref{tab:ri_IN-1K}-\ref{tab:ri_IN-21K-P}).
Linear Probing consistently achieves the best robustness improvement using LAION-2B pretrained models~(Table~\ref{tab:ri_LAION-2B}) and OpenAI CLIP models (Table~\ref{tab:ri_OpenAI}).

\input{method_first_tables/ri_IN-1K}
\input{method_first_tables/ri_SAM}
\input{method_first_tables/ri_IN-21K_wo_augreg}
\input{method_first_tables/ri_IN-21K}
\input{method_first_tables/ri_IN-21K-P}
\input{method_first_tables/ri_LAION-2B}
\input{method_first_tables/ri_openai}

\begin{figure*}[t!]
    \centering
    \includegraphics[width=0.8\linewidth]{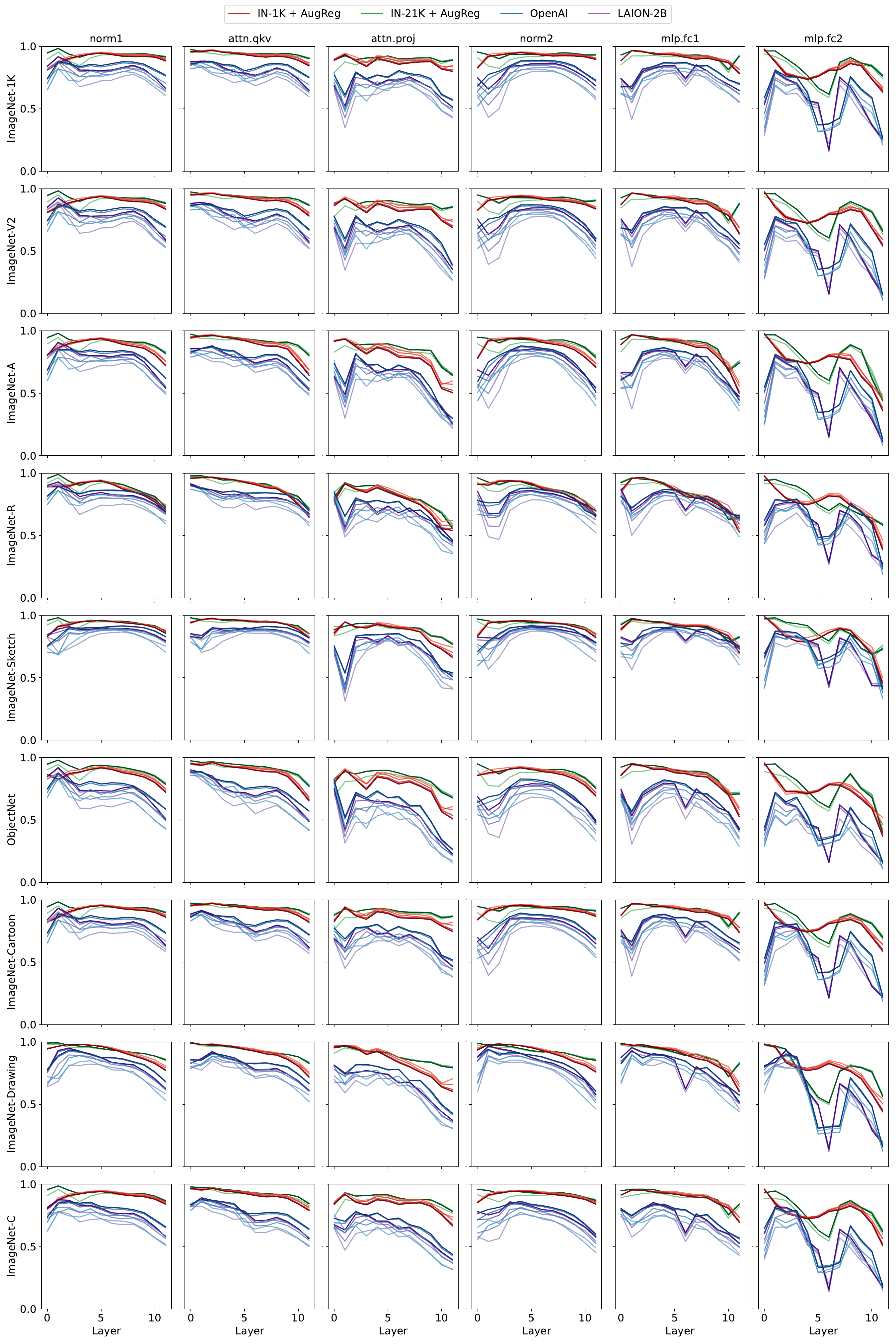}
\caption{
\textbf{Representational shifts while fine-tuning on ImageNet-R, analyzed by Centered Kernel Alignment (CKA), in ViT-B/16 models pretrained on various datasets.}
Rows indicate datasets where CKA is measured and column indicates different part of transformer demonstrated in Figure~\ref{fig:transformer}.
Darker and higher-contrast shades denote later epochs in the fine-tuning process.
ImageNet-R (downstream), ImageNet-1K (pretraining), and ImageNet-A (OOD dataset) are used as evaluation datasets.
Models pretrained on OpenAI and LAION-2B show distinct CKA patterns across layers compared to others.
}
\label{fig:cka_each}
\end{figure*}

\clearpage

\subsection{Trade-off between Downstream Accuracy and OOD Robustness}\label{supp:pareto}

\begin{figure*}[h!]
    \centering
    \includegraphics[width=0.9\linewidth]{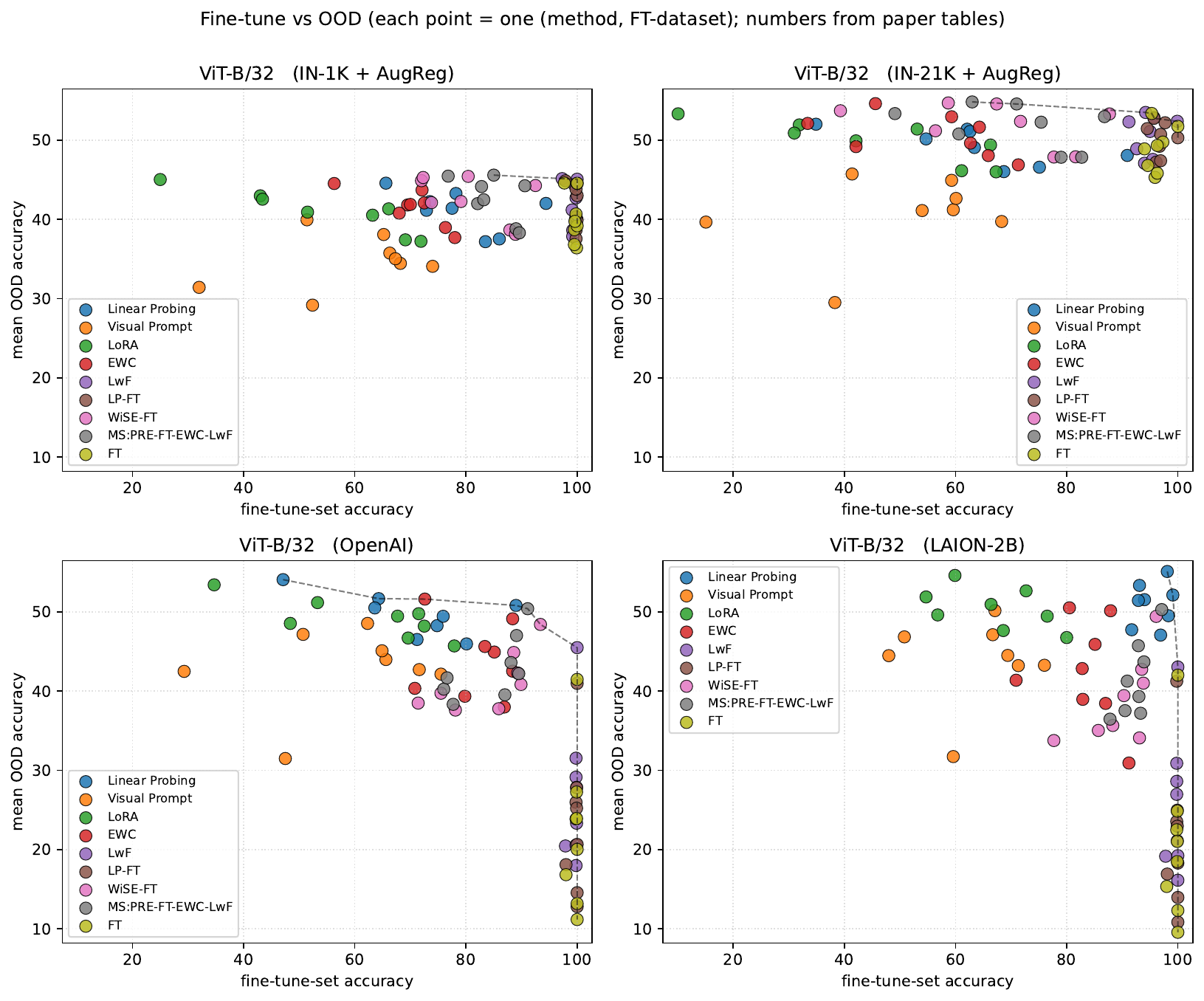}
\caption{
\textbf{Training accuracy versus OOD robustness across pretraining datasets (ViT-B/32).}
Each panel corresponds to one pretraining dataset.
Each point is one fine-tuned model, i.e., a (fine-tuning method, fine-tuning dataset) pair.
The $x$-axis shows the training accuracy on the fine-tuning dataset and the $y$-axis shows the mean accuracy on the seven held-out OOD datasets.
The dashed line indicates the Pareto envelope over all points.
Vanilla fine-tuning (FT) lies on the envelope for ImageNet-pretrained models but falls below it for CLIP-pretrained models (OpenAI, LAION-2B), where linear probing and weight-averaging methods span the envelope.
}
\label{fig:pareto}
\end{figure*}

Figure~\ref{fig:pareto} visualizes the trade-off between downstream accuracy and OOD robustness for each pretraining dataset.
Vanilla fine-tuning lies on the Pareto envelope for the ImageNet-pretrained models but falls below it for the CLIP-pretrained models (OpenAI and LAION-2B), whose envelope is instead spanned by linear probing and weight-averaging methods.

%% file: sections/7_supp.tex

\section{Analysis of Representation Shift via Centered Kernel Alignment (CKA)}
As shown in Figures~\ref{fig:cka_inv2}–\ref{fig:cka_inc}, models pretrained on LAION and OpenAI datasets generally exhibit lower CKA values compared to the other models. However, the gap between these two groups is relatively small when ImageNet-C is used for fine-tuning dataset. In any fine-tuning and evaluation dataset pair, sixth transformer's \textsf{mlp.fc2} exhibits significant drop.

\begin{table}[h!]
\centering
\caption{Reference for the Figures regarding Centered Kernel Alignment.}
\label{tab:figure_index}
\begin{tabular}{c|c}
\toprule
Fine-Tuning Dataset & Figure Index \\
\midrule
ImageNet-V2 & Figure~\ref{fig:cka_inv2} \\
ImageNet-A & Figure~\ref{fig:cka_ina} \\
ImageNet-R & Figure~\ref{fig:cka_inr} \\
ImageNet-Sketch & Figure~\ref{fig:cka_sketch} \\
ObjectNet & Figure~\ref{fig:cka_obj} \\
ImageNet-Cartoon & Figure~\ref{fig:cka_cartoon} \\
ImageNet-Drawing & Figure~\ref{fig:cka_drawing} \\
ImageNet-C & Figure~\ref{fig:cka_inc} \\
\bottomrule
\end{tabular}
\end{table}

\section{Accuracy of Using Various Pre-Trained Models on Each OOD Datasets and Each Corruption in ImageNet-C}

Table~\ref{tab:table_index} summarizes the Table indices for the accuracy on each OOD (out-of-distribution) dataset (Table~\ref{tab:main_base_16_in1k} and Tables~\ref{tab:main_base_16_sam}-\ref{tab:main_resnet50_in1k}) and ImageNet-C (Tables~\ref{tab:inc_base_16_in1k}-\ref{tab:inc_resnet50_in1k}) after fine-tuning on various datasets.
Each pre-trained and fine-tuned model is evaluated on ImageNet-C with 15 corruptions at severity levels ranging from 1 to 5.
Following the original ImageNet-C benchmark~\cite{hendrycks2019benchmarking}, we average the performance over the different severity levels. 
However, for consistency with other datasets, we report the results as accuracy rather than error.

\begin{table}[h!]
\centering
\caption{Reference for the tables showing accuracy of pre-trained models on OOD datasets (left) and ImageNet-C corruptions (right).}
\label{tab:table_index}
\begin{tabular}{c|c|c|c}
\toprule
Architecture&	$\Dpre$ & Accuracy on OOD datasets & Accuracy on ImageNet-C \\
\midrule
\multirow{7}{*}{ViT-B/16} & IN-1K + AugReg & Table~\ref{tab:main_base_16_in1k} & Table~\ref{tab:inc_base_16_in1k} \\
 & IN-1K + SAM & Table~\ref{tab:main_base_16_sam} & Table~\ref{tab:inc_base_16_sam} \\
 & IN-21K&  Table~\ref{tab:main_base_16_orig} & Table~\ref{tab:inc_base_16_orig} \\
 & IN-21K-P & Table~\ref{tab:main_base_16_in21kp} & Table~\ref{tab:inc_base_16_in21kp} \\
 & IN-21K + AugReg& Table~\ref{tab:main_base_16_imagenet} & Table~\ref{tab:inc_base_16_imagenet} \\
 & LAION-2B& Table~\ref{tab:main_base_16_laion} & Table~\ref{tab:inc_base_16_laion} \\
  & OpenAI & Table~\ref{tab:main_base_16_openai} & Table~\ref{tab:inc_base_16_openai} \\
\midrule
\multirow{4}{*}{ViT-B/32} & IN-1K + AugReg & 	Table~\ref{tab:main_base_32_in1k} & Table~\ref{tab:inc_base_32_in1k} \\
 & IN-21K  + AugReg	&	Table~\ref{tab:main_base_32_imagenet} & Table~\ref{tab:inc_base_32_imagenet} \\
 & LAION-2B& Table~\ref{tab:main_base_32_laion} & Table~\ref{tab:inc_base_32_laion} \\
  & OpenAI & Table~\ref{tab:main_base_32_openai} & Table~\ref{tab:inc_base_32_openai} \\
\midrule
\multirow{2}{*}{ViT-S/16}		 & IN-1K  + AugReg& Table~\ref{tab:main_small_16_in1k} & Table~\ref{tab:inc_small_16_in1k} \\
 & IN-21K  + AugReg& Table~\ref{tab:main_small_16_imagenet} & Table~\ref{tab:inc_small_16_imagenet} \\
\midrule
ViT-S/32 & IN-21K  + AugReg& Table~\ref{tab:main_small_32_imagenet} & Table~\ref{tab:inc_small_32_imagenet} \\
\midrule
\multirow{1}{*}{ViT-L/16}& IN-21K  + AugReg& Table~\ref{tab:main_large_16_imagenet} & Table~\ref{tab:inc_large_16_imagenet} \\
\midrule
\multirow{1}{*}{ResNet-18} & IN-1K  & Table~\ref{tab:main_resnet18_in1k} & Table~\ref{tab:inc_resnet18_in1k} \\
\midrule
\multirow{1}{*}{ResNet-50} & IN-1K &  Table~\ref{tab:main_resnet50_in1k} & Table~\ref{tab:inc_resnet50_in1k} \\
\bottomrule
\end{tabular}
\end{table}

\renewcommand{\thefigure}{S\arabic{figure}}
\setcounter{figure}{0}  

\begin{figure*}[t!]
    \centering
    \includegraphics[width=0.8\linewidth]{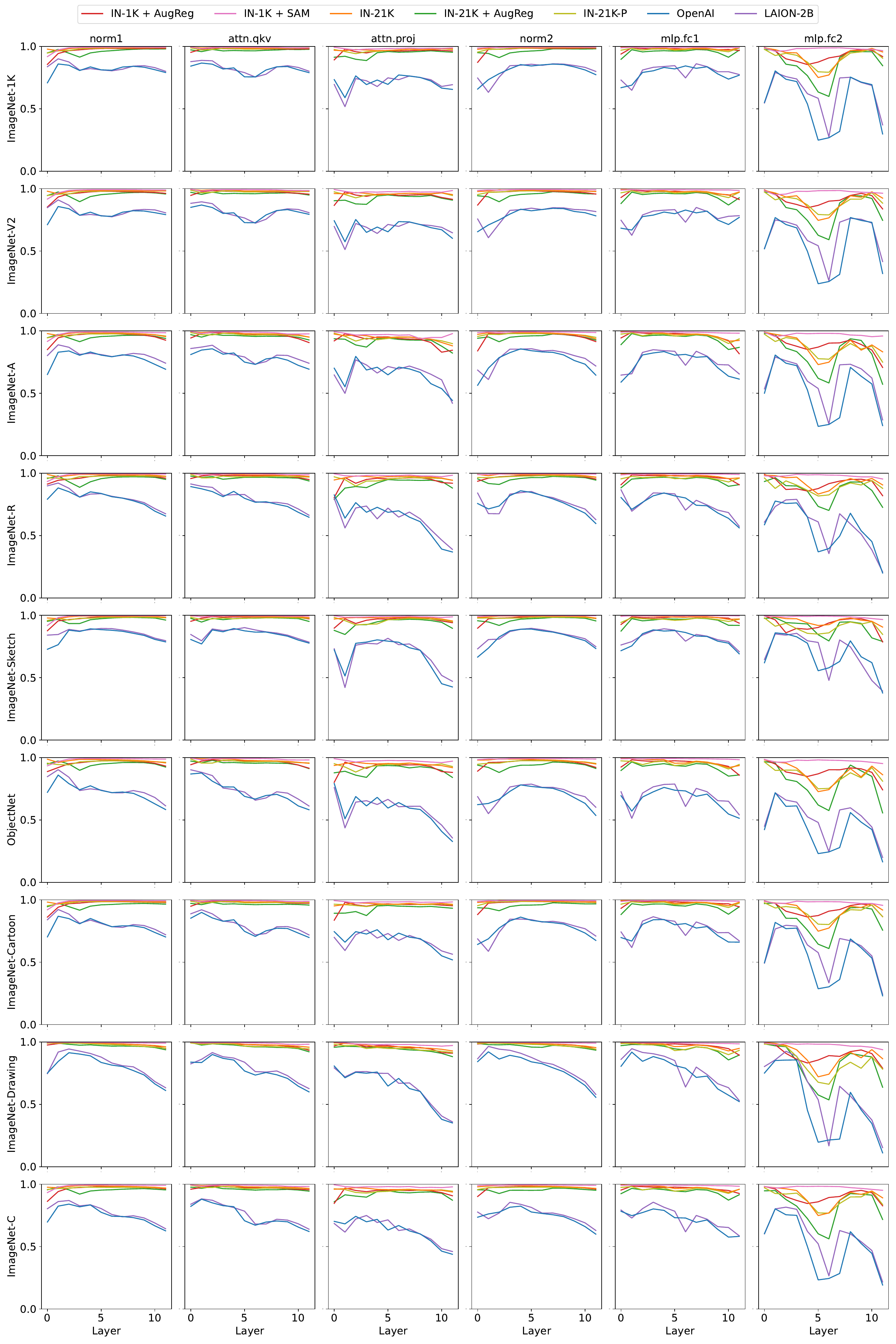}
\caption{
\textbf{Representational shifts from fine-tuning on ImageNet-V2, analyzed by Centered Kernel Alignment (CKA), in ViT-B/16 models pretrained on various datasets.}
Rows indicate datasets where CKA is measured, and column indicates different parts of the transformer demonstrated in Figure~9 in the main paper.
Models pretrained on OpenAI and LAION-2B show distinct CKA patterns across layers compared to others.
}
\label{fig:cka_inv2}
\end{figure*}

\begin{figure*}[t!]
    \centering
    \includegraphics[width=0.8\linewidth]{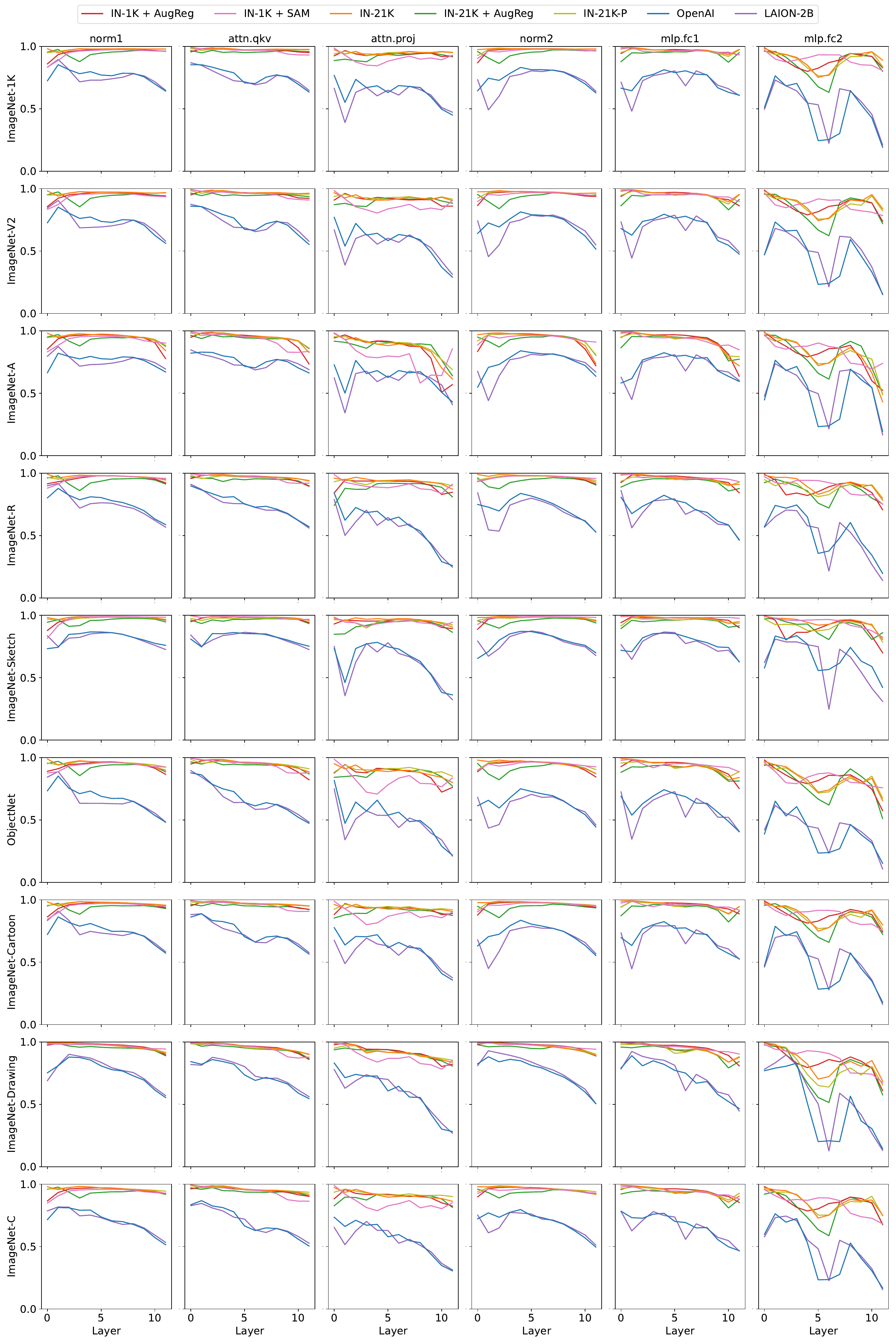}
\caption{
\textbf{Representational shifts from fine-tuning on ImageNet-A, analyzed by Centered Kernel Alignment (CKA), in ViT-B/16 models pretrained on various datasets.}
Rows indicate datasets where CKA is measured, and column indicates different parts of the transformer demonstrated in Figure~9 in the main paper.
Models pretrained on OpenAI and LAION-2B show distinct CKA patterns across layers compared to others.
}
\label{fig:cka_ina}
\end{figure*}

\begin{figure*}[t!]
    \centering
    \includegraphics[width=0.8\linewidth]{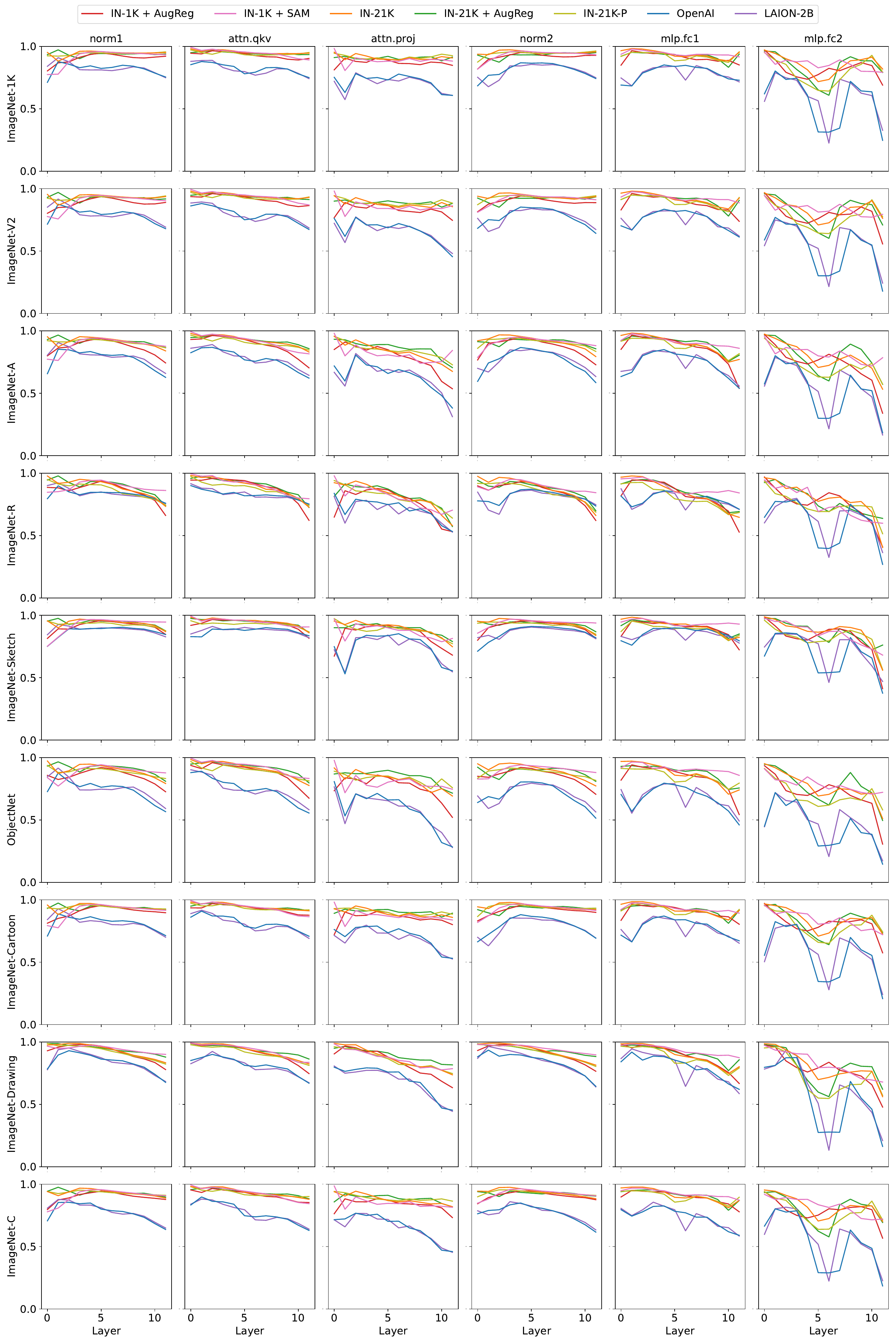}
\caption{
\textbf{Representational shifts from fine-tuning on ImageNet-R, analyzed by Centered Kernel Alignment (CKA), in ViT-B/16 models pretrained on various datasets.}
Rows indicate datasets where CKA is measured, and column indicates different parts of the transformer demonstrated in Figure~9 in the main paper.
Models pretrained on OpenAI and LAION-2B show distinct CKA patterns across layers compared to others.
}
\label{fig:cka_inr}
\end{figure*}

\begin{figure*}[t!]
    \centering
    \includegraphics[width=0.8\linewidth]{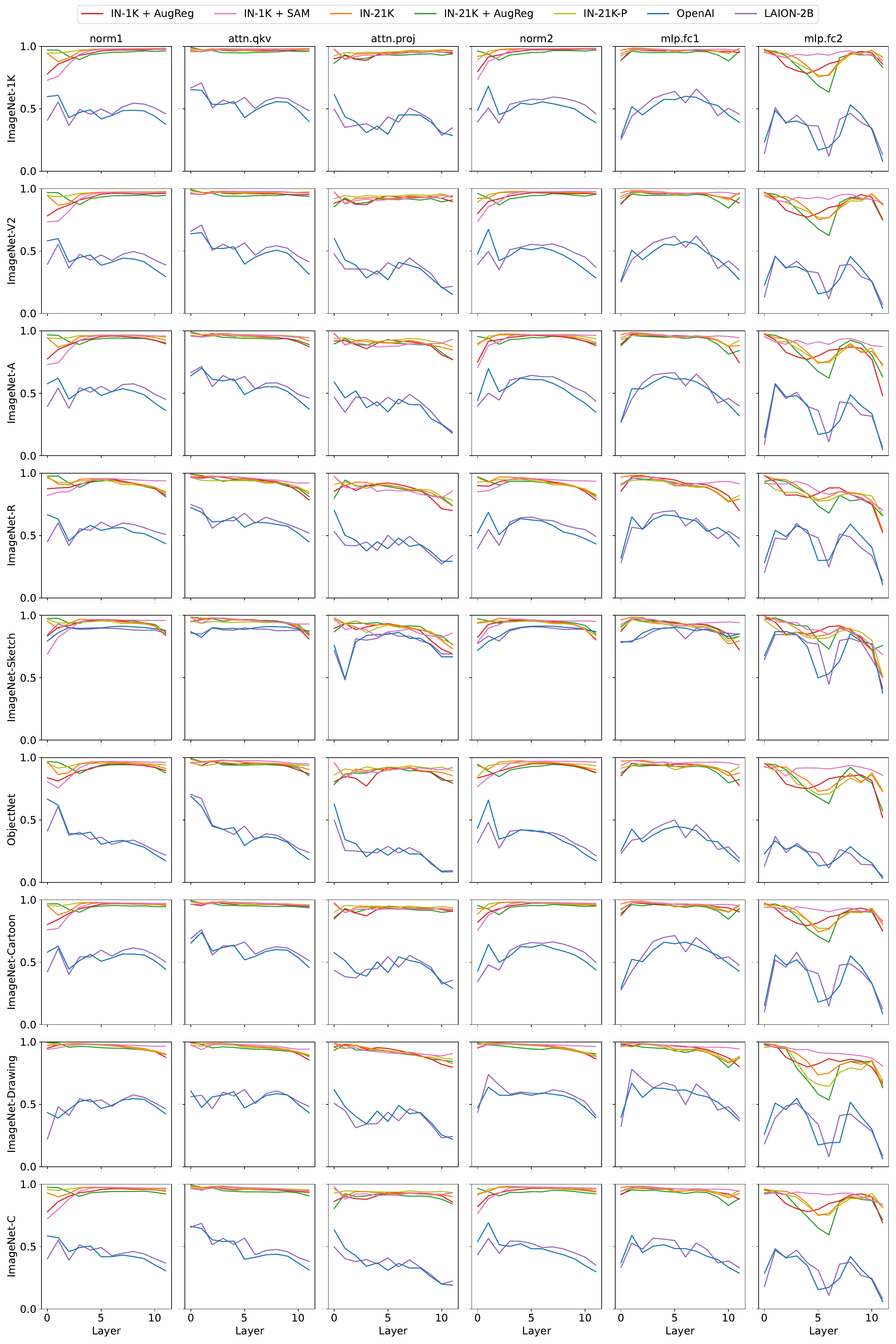}
\caption{
\textbf{Representational shifts from fine-tuning on ImageNet-Sketch, analyzed by Centered Kernel Alignment (CKA), in ViT-B/16 models pretrained on various datasets.}
Rows indicate datasets where CKA is measured, and column indicates different parts of the transformer demonstrated in Figure~9 in the main paper.
Models pretrained on OpenAI and LAION-2B show distinct CKA patterns across layers compared to others.
}
\label{fig:cka_sketch}
\end{figure*}

\begin{figure*}[t!]
    \centering
    \includegraphics[width=0.8\linewidth]{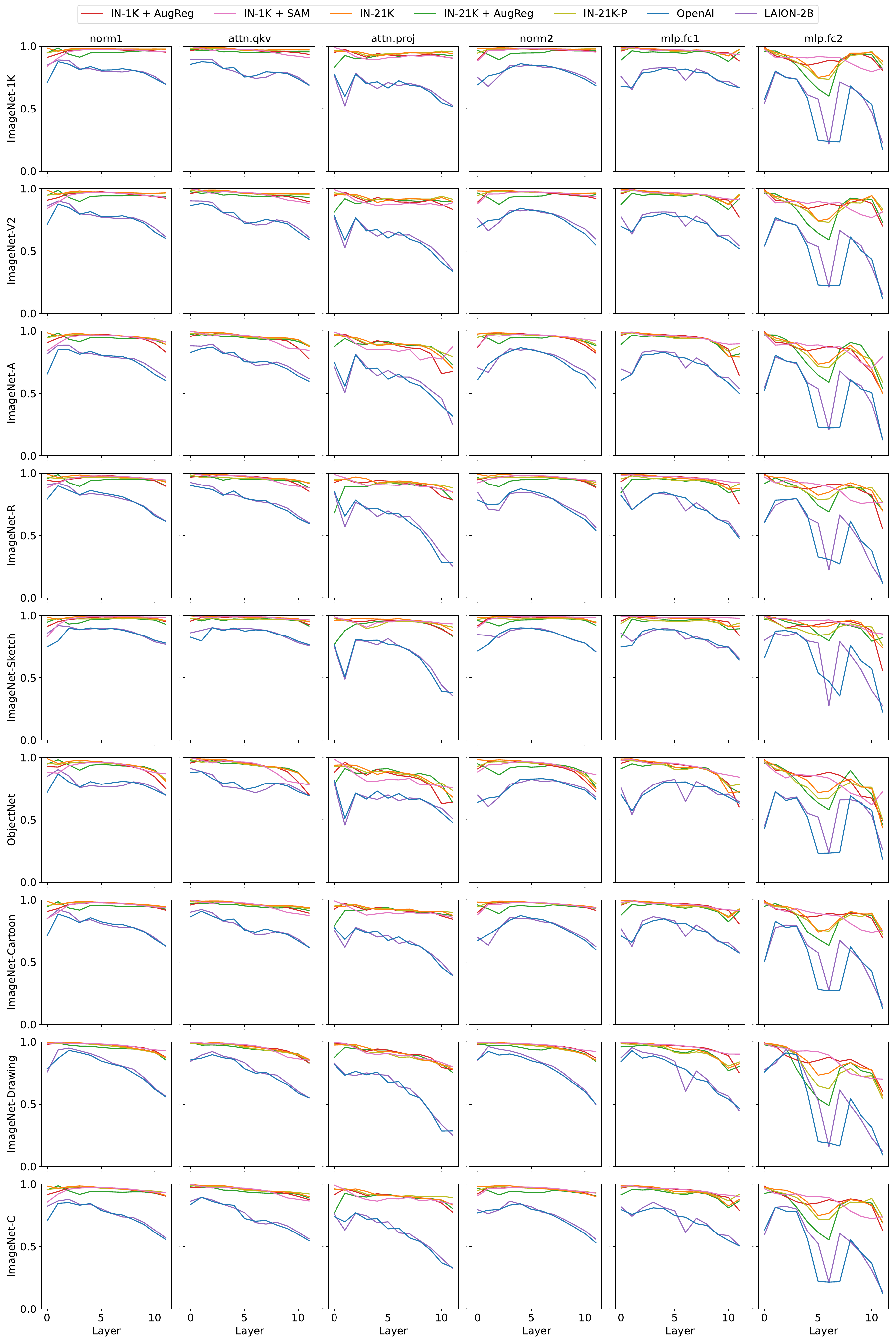}
\caption{
\textbf{Representational shifts from fine-tuning on ObjectNet, analyzed by Centered Kernel Alignment (CKA), in ViT-B/16 models pretrained on various datasets.}
Rows indicate datasets where CKA is measured, and column indicates different parts of the transformer demonstrated in Figure~9 in the main paper.
Models pretrained on OpenAI and LAION-2B show distinct CKA patterns across layers compared to others.
}
\label{fig:cka_obj}
\end{figure*}

\begin{figure*}[t!]
    \centering
    \includegraphics[width=0.8\linewidth]{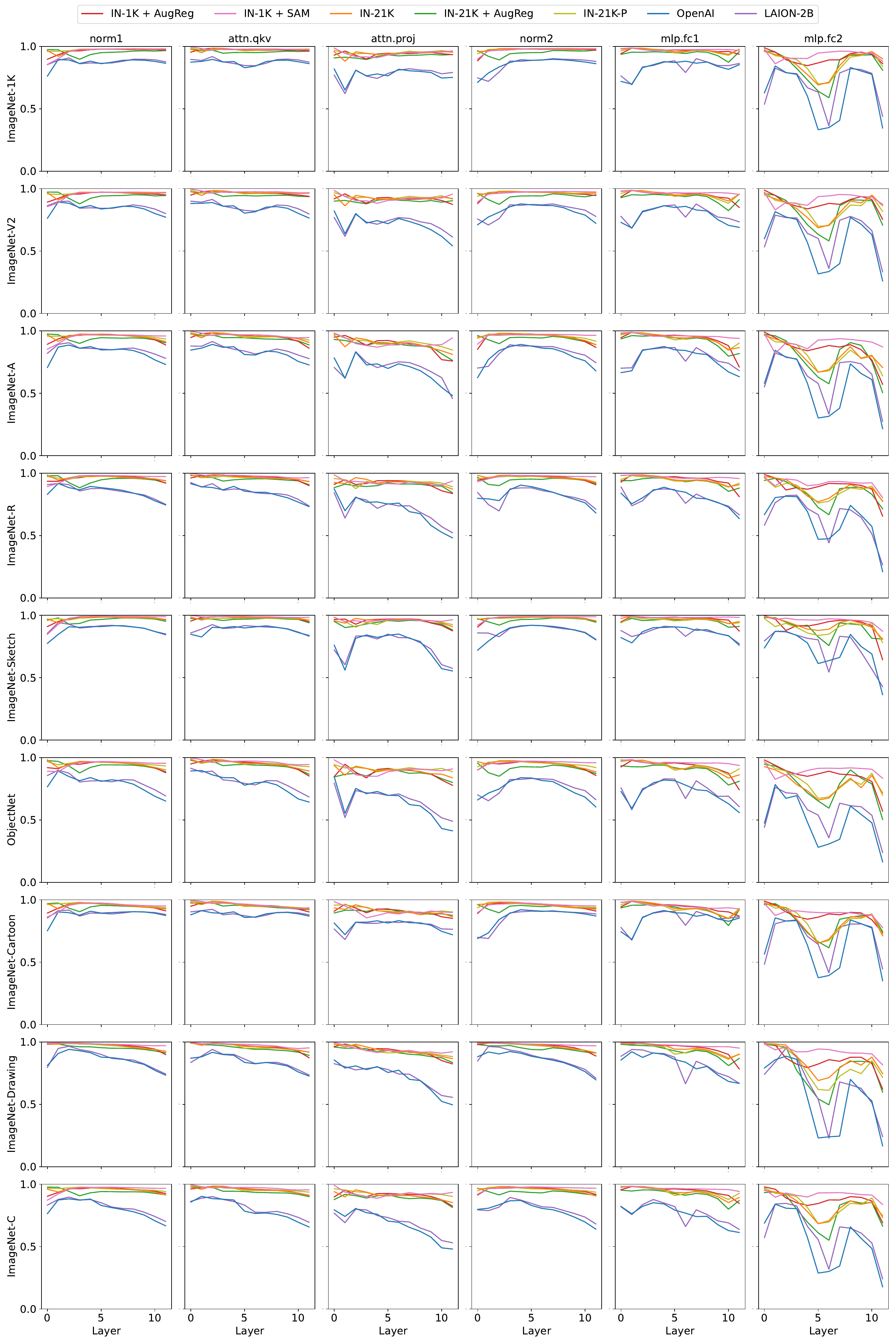}
\caption{
\textbf{Representational shifts from fine-tuning on ImageNet-Cartoon, analyzed by Centered Kernel Alignment (CKA), in ViT-B/16 models pretrained on various datasets.}
Rows indicate datasets where CKA is measured, and column indicates different parts of the transformer demonstrated in Figure~9 in the main paper.
Models pretrained on OpenAI and LAION-2B show distinct CKA patterns across layers compared to others.
}
\label{fig:cka_cartoon}
\end{figure*}

\begin{figure*}[t!]
    \centering
    \includegraphics[width=0.8\linewidth]{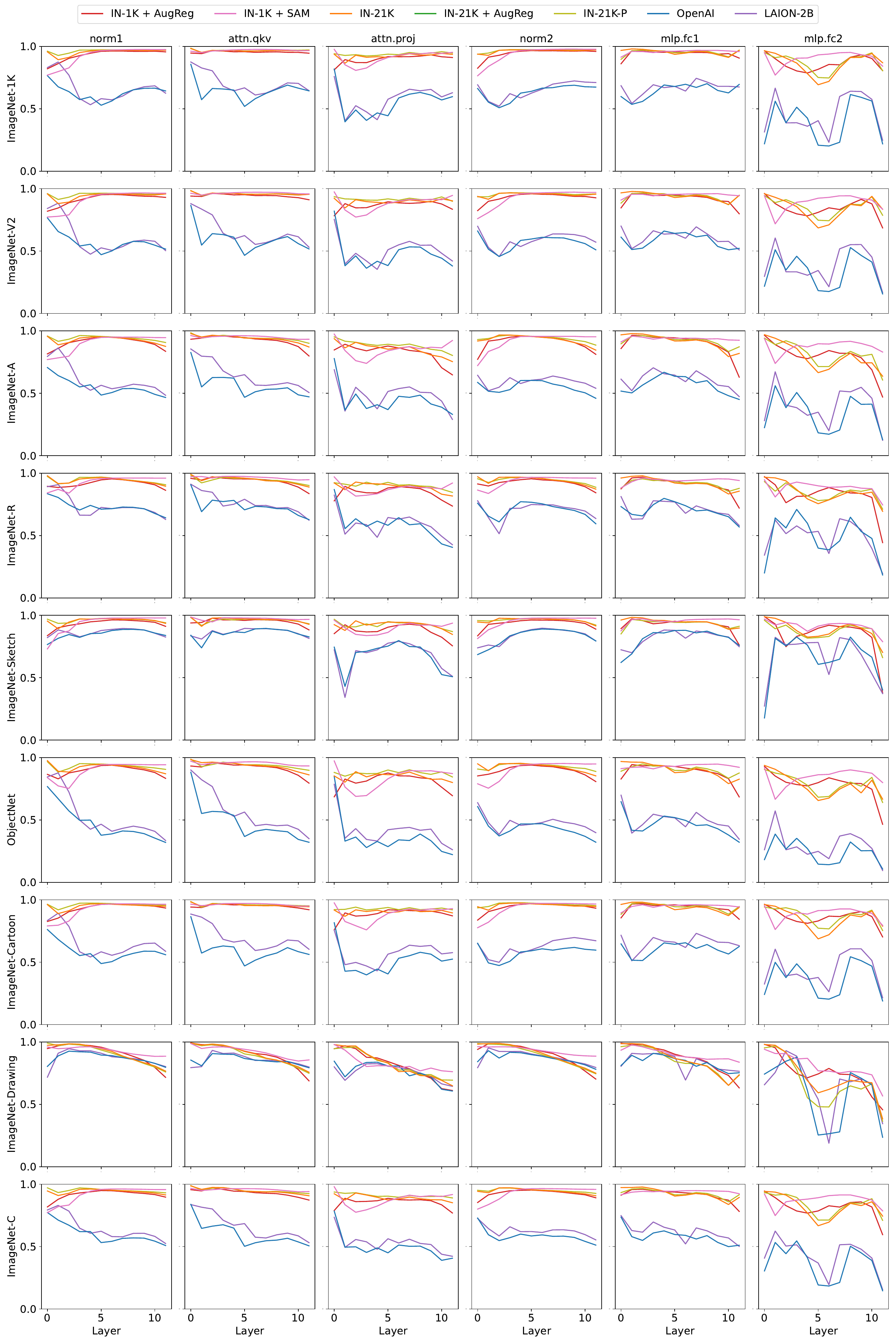}
\caption{
\textbf{Representational shifts from fine-tuning on ImageNet-Drawing, analyzed by Centered Kernel Alignment (CKA), in ViT-B/16 models pretrained on various datasets.}
Rows indicate datasets where CKA is measured, and column indicates different parts of the transformer demonstrated in Figure~9 in the main paper.
Models pretrained on OpenAI and LAION-2B show distinct CKA patterns across layers compared to others.
}
\label{fig:cka_drawing}
\end{figure*}
\begin{figure*}[t!]
    \centering
    \includegraphics[width=0.8\linewidth]{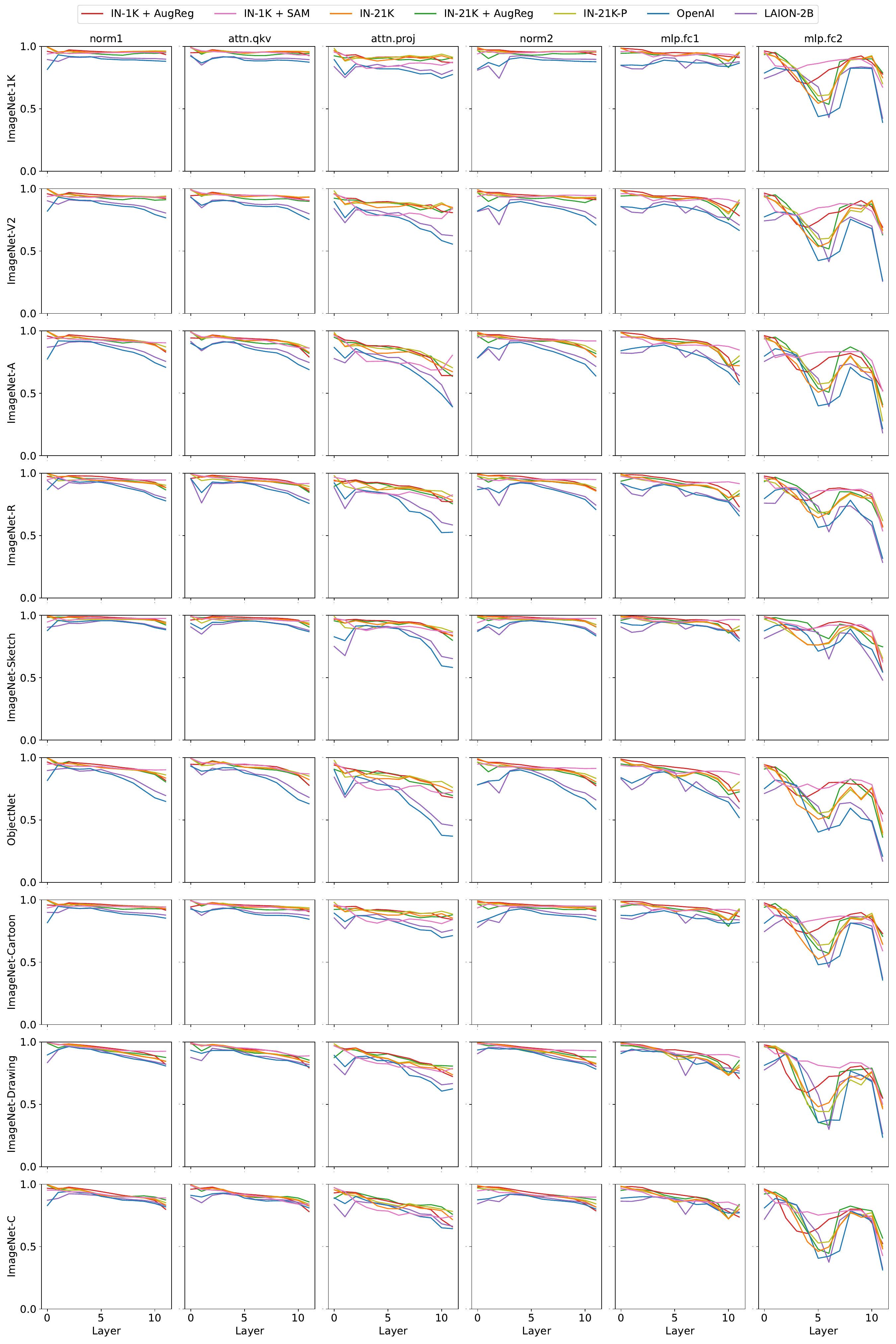}
\caption{
\textbf{Representational shifts from fine-tuning on ImageNet-C, analyzed by Centered Kernel Alignment (CKA), in ViT-B/16 models pretrained on various datasets.}
Rows indicate datasets where CKA is measured, and column indicates different parts of the transformer demonstrated in Figure~9 in the main paper.
Models pretrained on OpenAI and LAION-2B show distinct CKA patterns across layers compared to others.
}
\label{fig:cka_inc}
\end{figure*}

\renewcommand{\thetable}{S\arabic{table}}
\setcounter{table}{0}  


\input{method_first_tables/main_base_16_in1k}
\input{method_first_tables/main_base_16_sam}
\input{method_first_tables/main_base_16_orig}

\input{method_first_tables/main_base_16_in21kp}
\input{method_first_tables/main_base_16_imagenet}
\input{method_first_tables/main_base_16_laion}
\input{method_first_tables/main_base_16_openai}

\input{method_first_tables/main_base_32_in1k}
\input{method_first_tables/main_base_32_imagenet}
\input{method_first_tables/main_base_32_laion}
\input{method_first_tables/main_base_32_openai}
\input{method_first_tables/main_small_16_in1k}
\input{method_first_tables/main_small_16_imagenet}
\input{method_first_tables/main_small_32_imagenet}
\input{method_first_tables/main_large_16_imagenet}
\input{method_first_tables/main_resnet18_in1k}
\input{method_first_tables/main_resnet50_in1k}
\clearpage

\input{method_first_tables/inc_base_16_in1k}
\input{method_first_tables/inc_base_16_sam}
\input{method_first_tables/inc_base_16_orig}
\input{method_first_tables/inc_base_16_in21kp}

\clearpage
\input{method_first_tables/inc_base_16_imagenet}
\input{method_first_tables/inc_base_16_laion}
\input{method_first_tables/inc_base_16_openai}

\input{method_first_tables/inc_base_32_in1k}
\input{method_first_tables/inc_base_32_sam}
\input{method_first_tables/inc_base_32_imagenet}
\input{method_first_tables/inc_base_32_laion}
\input{method_first_tables/inc_base_32_openai}

\input{method_first_tables/inc_small_16_in1k}
\input{method_first_tables/inc_small_16_imagenet}
\input{method_first_tables/inc_small_32_imagenet}
\input{method_first_tables/inc_large_16_imagenet}

\input{method_first_tables/inc_resnet18_in1k}
\input{method_first_tables/inc_resnet50_in1k}